\documentclass[10pt,twocolumn]{article} 
\usepackage{simpleConference}
\usepackage{times}
\usepackage{amssymb}
\usepackage{url,hyperref}

\usepackage{adjustbox}
\usepackage{amsmath,amssymb,amsfonts}
\usepackage{cite}
\usepackage{bbm}
\usepackage{algorithmic}
\usepackage{graphicx}
\usepackage{textcomp}
\usepackage{hyperref}
\hypersetup{
    colorlinks=true,
    linkcolor=blue,
    filecolor=magenta,      
    urlcolor=cyan,
    pdftitle={Overleaf Example},
    pdfpagemode=FullScreen,
    }
\usepackage{authblk}
\usepackage{subcaption}
\usepackage{float}

\usepackage[para,online,flushleft]{threeparttable}
\usepackage{multirow}

\begin{document}

\title{A graph-transformer for whole slide image classification}

\author[1,2]{Yi Zheng}
\author[2]{Rushin H. Gindra}
\author[2]{Emily J. Green}
\author[3]{Eric J. Burks}
\author[1]{Margrit Betke}
\author[2]{Jennifer E. Beane}
\author[2,4]{Vijaya B. Kolachalama}
\affil[1]{Department of Computer Science, Boston University}
\affil[2]{Department of Medicine, Boston University School of Medicine}
\affil[3]{Department of Pathology \& Laboratory Medicine, Boston University School of Medicine}
\affil[4]{Department of Computer Science and Faculty of Computing \& Data Sciences, Boston University}

\maketitle
\thispagestyle{empty}

\begin{abstract}
The material in this template is an edited \& \LaTeX\--ified Deep learning is a powerful tool for whole slide image (WSI) analysis. Typically, when performing supervised deep learning, a WSI is divided into small patches, trained and the outcomes are aggregated to estimate disease grade. However, patch-based methods introduce label noise during training by assuming that each patch is independent with the same label as the WSI and neglect overall WSI-level information that is significant in disease grading. Here we present a Graph-Transformer (GT) that fuses a graph-based representation of an WSI and a vision transformer for processing pathology images, called GTP, to predict disease grade. We selected $4,818$ WSIs from the Clinical Proteomic Tumor Analysis Consortium (CPTAC), the National Lung Screening Trial (NLST), and The Cancer Genome Atlas (TCGA), and used GTP to distinguish adenocarcinoma (LUAD) and squamous cell carcinoma (LSCC) from adjacent non-cancerous tissue (normal). First, using NLST data, we developed a contrastive learning framework to generate a feature extractor. This allowed us to compute feature vectors of individual WSI patches, which were used to represent the nodes of the graph followed by construction of the GTP framework. Our model trained on the CPTAC data achieved consistently high performance on three-label classification (normal versus LUAD versus LSCC: mean accuracy$= 91.2$ $\pm$ $2.5\%$) based on five-fold cross-validation, and mean accuracy $= 82.3$ $\pm$ $1.0\%$ on external test data (TCGA). We also introduced a graph-based saliency mapping technique, called GraphCAM, that can identify regions that are highly associated with the class label. Our findings demonstrate GTP as an interpretable and effective deep learning framework for WSI-level classification.
\end{abstract}

\section{Introduction}
\label{sec:introduction}
Computational pathology \cite{arpa2014,arpa2015,path5331,FUCHS2011515}, which entails the analysis of digitized pathology slides, is gaining increased attention over the past few years. The sheer size of a single whole slide image (WSI) typically can exceed a gigabyte, so traditional image analysis routines may not be able to fully process all this data in an efficient fashion. Modern machine learning methods such as deep learning have allowed us to make great progress in terms of analyzing WSIs including disease classification \cite{8822590}, tissue segmentation \cite{WANG20191686}, mutation prediction \cite{nm2018}, and spatial profiling of immune infiltration \cite{Saltz2018SpatialOA}. Most of these methods rely on systematic breakdown of WSIs into image patches, followed by development of deep neural networks at patch-level and integration of outcomes on these patches to create overall WSI-level estimates \cite{7780635,Wei2019}. While patch-based approaches catalyzed research in the field, the community has begun to appreciate the conditions in which they confer benefit and in those where they cannot fully capture the underlying pathology. For example, methods focused on identifying the presence or absence of a tumor on an WSI can be developed on patches using computationally efficient techniques such as multiple instance learning \cite{das9269335}. On the other hand, if the goal is to identify the entire tumor region or capture the connectivity of the tumor microenvironment characterizing the stage of disease, then it becomes important to assess both regional and WSI-level information. There are several other scenarios where both the patch- and WSI-level features need to be identified to assess the pathology \cite{ZHENG20211442}, and methods to perform such analysis are needed.

The success of patch-based deep learning methods can be attributed to the availability of pre-trained deep neural networks on natural images from public databases (i.e., ImageNet \cite{deng5206848}). Since there are millions of parameters in a typical deep neural network, \emph{de novo} training of this network requires access to a large set of pathology data, and such resources are not necessarily available at all locations. To address this bottleneck, researchers have leveraged transfer learning approaches that are pre-trained on ImageNet to accomplish various tasks. Recently, transformer architectures were applied directly to sequences of image patches for various classification tasks. Specifically, Vision Transformers (ViT) were shown to achieve excellent results compared to state-of-the-art convolutional networks while requiring substantially fewer computational resources for model training \cite{dosovitskiy2020vit}. Position embeddings were used in ViTs to retain spatial information and capture the association of different patches within the input image. The self-attention mechanism in ViT requires the calculation of pairwise similarity scores on all the patches, resulting in memory efficiency and a simple time complexity that is quadratic in the number of patches. Leveraging such approaches to perform pathology image analysis is not trivial because each WSI can contain thousands of patches. Additionally, some approximations are often made on these patches such as using the WSI-level label on each patch during training, which is not ideal in all scenarios as there is a need to process both the regional information as well as the WSI in its entirety to better understand the pathological correlates of disease. 

Similar to the regional and WSI-level examination, we argue that an expert pathologist's workflow also involves examination of the entire slide under the microscope using manual operations such as panning and zooming in and out of specific regions of interest to assess various aspects of disease at multiple scales. In the zoom-in assessment, the pathologists perform an in-depth, evaluation of regional manifestations of disease whereas, the zoom-out assessment involves obtaining a rational estimate of the overall disease on the entire WSI. Both these assessments are critical as the pathologist obtains a gestalt on various image features to comprehensively assess the disease \cite{ZHENG20211442}. 

\subsection{Related work}
Recent attempts to perform WSI-level analysis have shown promising results in terms of assessing the overall tissue microenvironment. In particular, graph-based approaches such as graph convolutional networks have gained a lot of traction due to their ability to represent the entire WSI and analyze patterns to predict various outcomes of interest. To learn hierarchical representation for graph embedding, recent approaches have proposed pooling strategies. For example, in AttPool \cite{huang2019attpool}, Huang and colleagues devised an attention pooling layer to select discriminative nodes and built the coarser graph based on calculated attention values. AttPool sufficiently leveraged the hierarchical representation and facilitated model learning on several graph classification benchmark datasets. Zhou and colleagues developed a cell-graph convolutional neural network on WSIs to predict the grade of colorectal cancer (CRC) \cite{ZhouGKSHR19}. In this work, the WSI was converted to a graph, where each nucleus was represented by a node and the cellular interactions were denoted as edges between these nodes to accurately predict CRC grade. Also, Adnan and colleagues developed a two-stage framework for WSI representation learning \cite{Adnan2020}, where patches were sampled based on color and a graph neural network was constructed to learn the inter-patch relationships to discriminate lung adenocarcinoma (LUAD) from lung squamous cell carcinoma (LSCC). In another recent work, Lu and team developed a graph representation of the cellular architecture on the entire WSI to predict the status of human epidermal growth factor receptor 2 and progesterone receptor \cite{lu9150693}. Their architecture attempted to create a bottom-up approach (i.e., nuclei- to WSI-level) to construct the graph, and in so doing, achieved a relatively efficient framework for analyzing the entire WSI. 

Several other researchers have leveraged graph-based approaches to process WSIs to address various questions focused on survival analysis \cite{li2018,di2020,chen2021,9667788}, prediction of lymph node metastasis \cite{Zhao_2020}, mutational prediction \cite{ding2020}, cell classification \cite{SHI2021105807}, and retrieval of relevant regions \cite{ZHENG2022102308}. With the objective of retaining the correlation among different patches within an WSI, Shao and colleagues leveraged the computationally efficient sampling approach of multiple instance learning and the self-attention mechanism of ViT in TransMIL\cite{transmil}. To deal with the memory and quadratic time complexity issue in ViT, they adopted the Nystrom Method\cite{Xiong2021NystrmformerAN}. They expressed the significance of retaining these correlations among instances by showing good performance for tumor classification in $3$ different datasets: CAMELYON16, TCGA-NSCLC and TCGA-RCC. Motivated by these advances, we submit that integration of computationally efficient approaches such as ViTs along with graphs can lead to more efficient representation learning approaches for the assessment of WSIs.

\subsection{Contributions}
The main contributions of this paper are summarized below:
\begin{itemize}
    \item We developed a graph-based vision transformer for digital pathology called GTP that leverages a graph representation of pathology images and the computational efficiency of transformer architectures to perform WSI-level analysis. To build the GTP framework, we constructed a graph convolutional network by embedding image patches in feature vectors using contrastive learning, followed by the application of a vision transformer to predict a WSI-level label.
    \item Using WSI and clinical data from three publicly available national cohorts ($4,818$ WSIs), we developed a model that could distinguish between normal, LUAD, and LSCC WSIs.
    \item We introduced graph-based class activation mapping (GraphCAM), a novel approach to generate WSI-level saliency maps that can identify image regions that are highly associated with the output class label. On a few WSIs, we also compared the performance of the GraphCAMs with pathologist-driven annotations and showed that our approach identifies important disease-related regions of interest.
    \item Over a series of ablation studies and sensitivity analyses, we showed that our GTP framework outperforms current state-of-the-art methods used for WSI classification.
\end{itemize}

\section{Materials and methods}

\subsection{Study population}
We obtained access to WSIs as well as demographic and clinical data of lung tumors (LUAD and LSCC) and normal tissue from the Clinical Proteomic Tumor Analysis Consortium (CPTAC), the National Lung Screening Trial (NLST) and The Cancer Genome Atlas (TCGA) (Table ~\ref{table:simDisimCoefNewDef}). CPTAC is a national effort to accelerate the understanding of the molecular basis of cancer through the application of large-scale proteome and genome analysis \cite{pr501254j}. NLST was a randomized controlled trial to determine whether screening for lung cancer with low-dose helical computed tomography reduces mortality from lung cancer in high-risk individuals relative to screening with chest radiography \cite{NEJMoa1102873}. TCGA is a landmark cancer genomics program, which molecularly characterized thousands of primary cancer and matched normal samples spanning $33$ cancer types \cite{tcga}.  

\begin{table}[H]
\caption{\textbf{Study population}. Whole slide images and corresponding clinical information from three distinct cohorts including the Clinical Proteomic Tumor Analysis Consortium (CPTAC), The Cancer Genome Atlas (TCGA) and the National Lung Screening Trial (NLST) were used.}
\begin{center}
\begin{threeparttable}
(a) CPTAC
\begin{tabular}{l|l}
\hline
    Description                            & Value \\ \hline
    Number of patients                     & 435    \\
    Number of WSIs           & 2071      \\
    Number of WSIs per class$^1$ & 719, 667, 685      \\
    Number of patches$^2$                  & 1277, [100-8478]       \\
    Age $^3$                               & 1, 4, 23, 80, 134, 89, 5, 99 \hspace{0.32cm} \\
    Gender $^4$                            & 235, 101, 99       \\
    Race $^5$            & 89, 5, 1, 1, 339      \\\hline
\end{tabular}
(b) TCGA
\begin{tabular}{l|l}
\hline
    Description                            & Value \\ \hline
    Number of patients                     & 996    \\
    Number of WSIs           & 2082      \\
    Number of WSIs per class$^1$ & 534, 808, 740      \\
    Number of patches$^2$                  & 469.0, [100-5491]      \\
    Age $^3$                               & 0, 2, 40, 149, 340, 366, 74, 24 \\
    Gender $^4$                            & 598, 398, 0      \\
    Race $^5$                 & 718, 84, 17, 1, 176      \\\hline
\end{tabular}
(c) NLST
\begin{tabular}{l|l}
\hline
    Description                            & Value \\ \hline
    Number of patients                     & 345    \\
    Number of WSIs           & 665      \\
    Number of WSIs per class$^1$ & 75, 378, 212      \\
    Number of patches$^2$                  & 2679.5, [110-7029]      \\
    Age $^3$                               & 0, 0, 0, 87, 201, 57, 0, 0 \hspace{0.57cm} \\
    Gender $^4$                            & 211,134, 0      \\
    Race $^5$                & 315, 14, 11, 1, 4      \\ \hline
\end{tabular}
\begin{tablenotes}
\item[1] Normal, LUAD, LSCC \quad
\item[2] Median, Range \\
\item[3] Binned: 20-29, 30-39, 40-49, 50-59, 60-69, 70-79, 80-89, Unknown \\
\item[4] Males, Females, Unknown \\
\item[5] White, Black or African-American, Asian, American Indian or Alaskan Native, Other/unknown \\
\end{tablenotes}
\end{threeparttable}
\end{center}
\label{table:simDisimCoefNewDef}
\vspace{-10mm}
\end{table}

\begin{figure*}[hbtp]
  \centering
  \begin{subfigure}{\linewidth}
    \centering
    \includegraphics[width=0.95\linewidth]{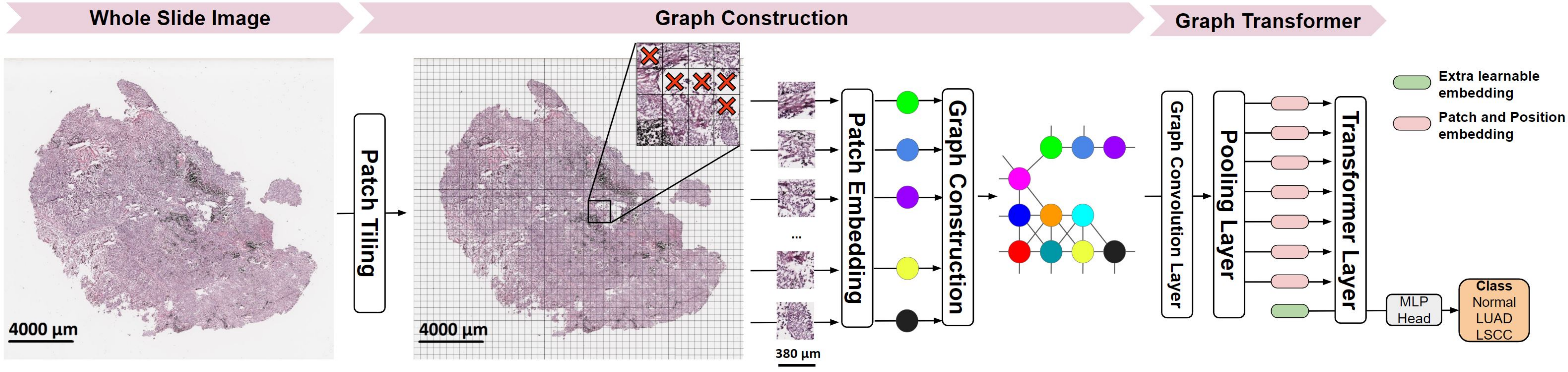}
    \caption[]%
            {{\small Schematic of the graph-transformer.}}
  \end{subfigure} 

  \begin{subfigure}{\linewidth}
    \centering
    \includegraphics[width=0.7\linewidth]{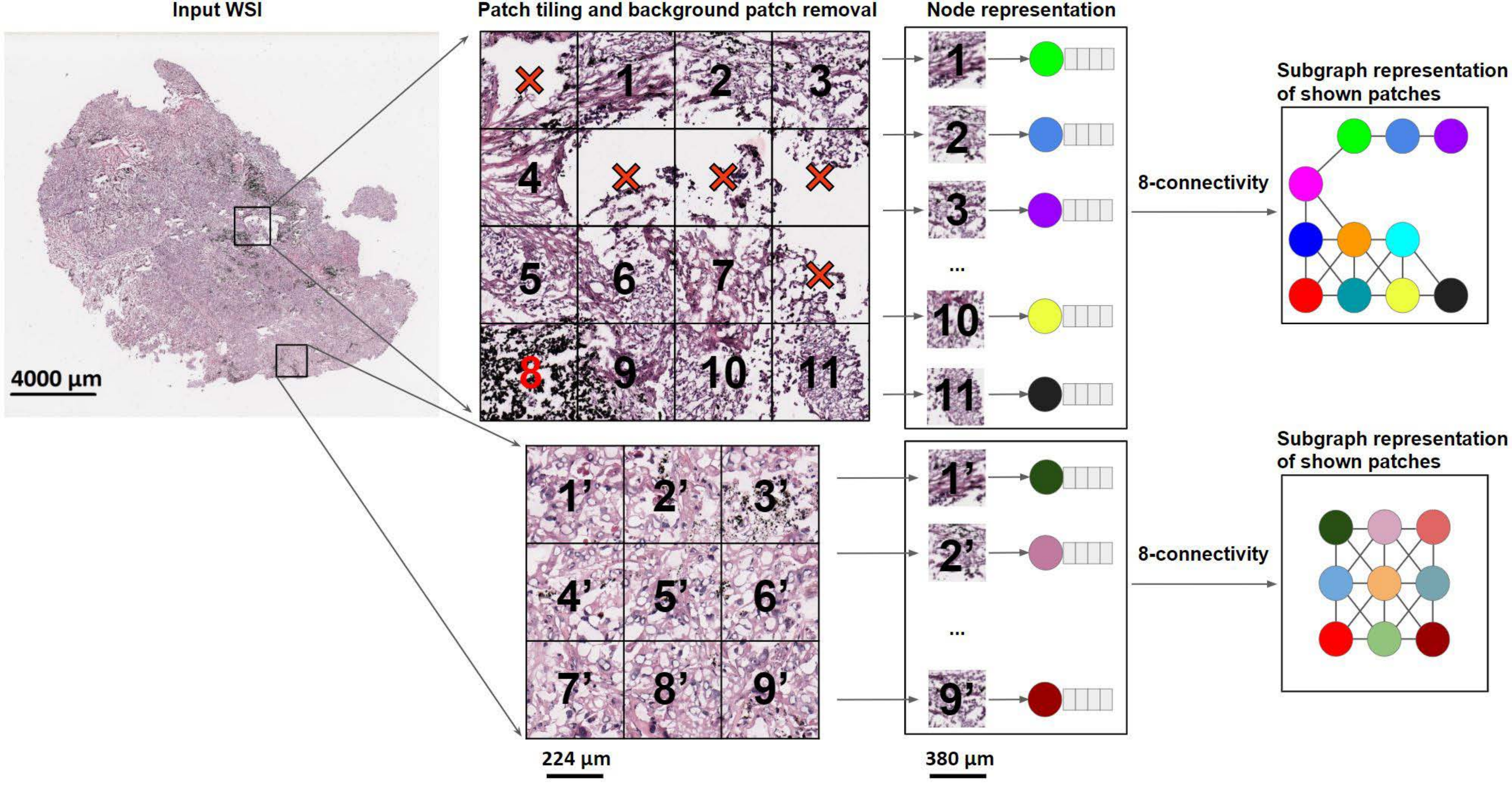}
    \caption[]%
            {{\small Node representation and connectivity information.}}    
  \end{subfigure}  

    \begin{subfigure}{\linewidth}
    \centering
    \includegraphics[width=0.75\linewidth]{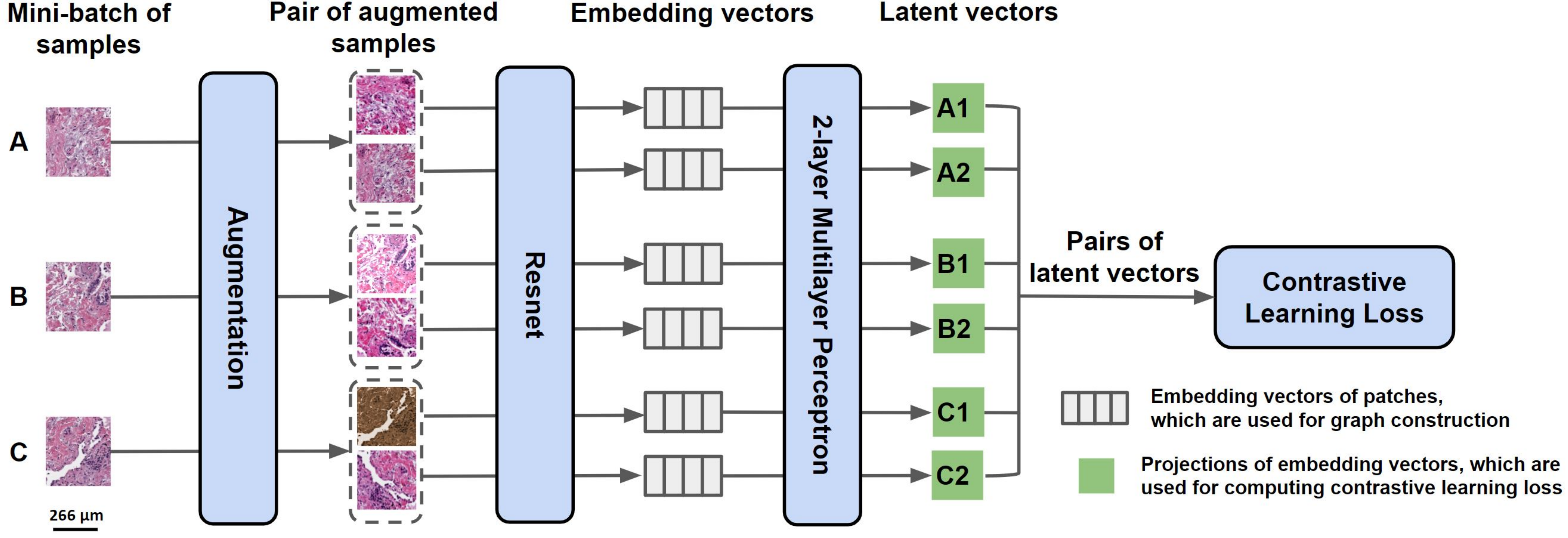}
    \caption[]%
            {{\small Feature generation and contrastive learning.}}    
  \end{subfigure}  
  \caption{\textbf{The GTP framework.} (a) Each whole slide image (WSI) was divided into patches. Patches that predominantly contained the background were removed, and the remaining patches were embedded in feature vectors by a contrastive learning-based patch embedding module. The feature vectors were then used to build the graph followed by a transformer that takes the graph as the input and predicts WSI-level class label. (b) Each selected patch was represented as a node and a graph was constructed on the entire WSI using the nodes with an 8-node adjacency matrix. Here, two sets of patches of an WSI and their corresponding subgraphs are shown. The subgraphs are connected within the graph representing the entire WSI. (c) We applied two distinct augmentation functions, including random color distortions, random Gaussian blur, and random cropping followed by resizing back to the original size, on the same sample in a mini-batch. If the mini-batch size is $K$, then we ended up with $2\times K$ augmented observations in the mini-batch. The ResNet received an augmented image leading to an embedding vector as the output. Subsequently, a projection head was applied to the embedding vector which produced the inputs to contrastive learning. The projection head is a multilayer perceptron (MLP) with 2 dense layers. In this example, we considered $K=3$ samples in a minibatch (A, B \& C). For sample A, the positive pairs are (A1, A2) and (A2, A1), and the negative pairs are (A1, B1), (A1, B2), (A1, C1), (A1, C2). All pairs were used for computing contrastive learning loss to train the Resnet. After training, we used the embedding vectors (straight from the ResNet) for constructing the graph.}  
\label{fig:overall-architecture}
\end{figure*}

\subsection{Graph-Transformer} 

Our proposed Graph-Transformer (GT) (Fig.~\ref{fig:overall-architecture} (a)) fuses a graph representation $G$ of an WSI (Fig.~\ref{fig:overall-architecture} (b)), and a transformer that can generate WSI-level predictions in a computationally efficient fashion. 
Let $G = (V, E)$ be an undirected graph where $V$ is the set of nodes representing the image patches and $E$ is the set of edges between the nodes in $V$ that represent whether two image patches are adjacent to each other. We denote the adjacency matrix of $G$ as $\mathcal{A} = [\mathcal{A}_{ij}]$ where $\mathcal{A}_{ij} = 1$ if there exists an edge $(v_i, v_j) \in E$ and $\mathcal{A}_{ij}=0$ otherwise. 
An image patch must be connected to other patches and can be surrounded by at most 8 adjacent patches, so the sum of each row or column of ${\mathcal{A}}$ is at least one and at most 8.
A graph can be associated with a node feature matrix $F$, $F \in {\rm I\!R}^{N \times D}$, where 
each row contains the $D$-dimensional feature vector computed for an image patch, i.e., node, and $N=|V|$.

Using all the pixels within each image patch as features can make model training computationally intractable. Instead, our framework applies a feature extractor to generate a vector containing features and uses it to define the information contained in an image patch, which is a node in the graph. This step reduces the node feature dimension from $W_{p}\times H_{p} \times C_{p}$ to $D$, where $W_{p}, H_{p},$ and $C_{p}$ are width, height, and channel of the image patch, and $D \times 1$ is the dimension of extracted feature vector. The expectation is that the derived feature vector provides an efficient representation of the node and also serves as a robust means by which to define a uniform representation of an image patch for graph-based classification.

As described above, current methods that have been developed at patch-level impose WSI-level labels on all the patches or use weakly supervised learning to extract feature vectors that are representative of the WSI. This strategy is not suitable for all scenarios, especially when learning the overall WSI-level information. We leveraged a strategy based on self-supervised contrastive learning \cite{chen2020simple}, to extract features from the WSIs. This framework enables robust representations that can be learned without the need for manual labels. Our approach involves using contrastive learning to train a convolutional neural network (CNN) that produces embedding representations by maximizing agreement between two differently augmented views of the same image patch via a contrastive loss in the latent space (Fig.~\ref{fig:overall-architecture}(C)). The training starts with tiling the WSIs from the training set into patches and randomly sampling a mini-batch of $K$ patches. Two different data augmentation operations are applied to each patch ($p$), resulting in two augmented patches ($p_i$ and $p_j$). The pair of two augmented patches from the same patch is denoted as a positive pair. For a mini-batch of $K$ patches, there are $2K$ augmented patches in total. Given a positive pair, the other $2K-1$ augmented patches are considered as negative samples. Subsequently, the CNN is used to extract representative embedding vectors ($f_i$, $f_j$) from each augmented patch ($p_i$, $p_j$). The embedding vectors are then mapped by a projection head to a latent space ($z_i$, $z_j$) where contrastive learning loss is applied. The contrastive learning loss function for a positive pair of augmented patches (i, j) is defined as:
    \begin{equation}
        l_{i,j}=-\log\frac{\exp(sim(\textbf{z}_i,\textbf{z}_j)/\tau)}{\sum^{2K}_{k=1}{\mathbbm{1}}_{[k\ne i]}\exp(sim(\textbf{z}_i,\textbf{z}_k)/\tau)},
    \end{equation}
where ${\mathbbm{1}}_{[k\ne i]} \in \{0,1\}$ is an indicator function evaluating to 1 if and only if $k \ne i$ and $\tau$ denotes a temperature parameter. Also, $sim(\textbf{u}, \textbf{v}) = \textbf{u}^{T} \textbf{v}/\lVert \textbf{u} \rVert \lVert \textbf{v} \rVert$ denotes the dot product between $L_2$ normalized \textbf{u} and \textbf{v} (i.e., cosine similarity). For model training, the patches were densely cropped without overlap and treated as individual images. The final loss was computed across all positive pairs, including both (i, j) and (j, i) in a mini-batch. After convergence, we kept the feature extractor and used it for our GTP model to compute the feature vectors of the patches from the WSIs. GTP uses these computed feature vectors as node features in the graph construction phase. Specifically, we obtained the node-specific feature matrix $F=[f_1; f_2; \dots; f_N]$, $F \in {\rm I\!R}^{N \times D}$, where $f_i$ is the D-dimensional embedding vector obtained from Resnet trained using contrastive learning and $N$ is the number of patches from one WSI. Note that $N$ is variable since different WSIs contain different numbers of patches. As a result, each node in $F$ corresponds to one patch of the WSI. We defined an edge between a pair of nodes in $F$ based on the spatial location of its corresponding patches on the WSI. If patch $i$ is a neighbor of patch $j$ on the WSI (Fig.~\ref{fig:overall-architecture}(B)), then GTP creates an edge between node $i$ and node $j$ as well as set $\mathcal{A}_{ij}=1$ and $\mathcal{A}_{ji}=1$, otherwise $\mathcal{A}_{ij}=0$ and $\mathcal{A}_{ji}=0$. GTP uses feature node matrix $F$ and adjacent matrix $\mathcal{A}$ to construct a graph to represent each WSI.

We implemented the graph convolutional (GC) layer, introduced by Kipf \& Welling \cite{kipf2017semisupervised}, to handle the graph-structured data. The GC layer operates message propagation and aggregation in the graph, and is defined as:
    \begin{subequations}\label{eq:2}
    \begin{align}
    \label{eq:2a}
    &  H_{\rm m+1}=ReLU(\Hat{\mathcal{A}}H_{\rm m}W_{\rm m}), \quad \quad m=1,2,..,M \\
    \label{eq:2c}
    & \Hat{\mathcal{A}}=\Tilde{D}^{-\frac{1}{2}}\Tilde{\mathcal{A}}\Tilde{D}^{-\frac{1}{2}} 
    \end{align}
    \end{subequations}
where $\Hat{\mathcal{A}}$ is the symmetric normalized adjacency matrix of $\mathcal{A}$ and $M$ is the number of GC layers. Here, $\Tilde{\mathcal{A}}=\mathcal{A}+I$ is the adjacency matrix with a self-loop added to each node, and $\Tilde{D}$ is a diagonal matrix where $\Tilde{D}_{ii}=\sum_{j}\Tilde{\mathcal{A}}_{ij}$. $H_{m}$ is the input of the $m$-th GC layer and $H_1$ is initialized with the node feature matrix $F$. Additionally, $W_{\rm m} \in {\rm I\!R}^{C_{\rm m}\times C_{\rm m+1}}$ is the matrix of learnable filters in the GC layer, where $C_{\rm m}$ is the dimension of the input and $C_{\rm m+1}$ is the dimension of the output.

The GC layer of GTP enables learning of node embeddings through propagating and aggregating needed information. However, it is not trivial for a model to learn hierarchical features that are crucial for graph representation and classification. To address this limitation, we introduced a transformer layer that selects the most significant nodes in the graph and aggregates information via the attention mechanism. Transformers use a Self-Attention (SA) mechanism to model the interactions between all tokens in a sequence \cite{vaswani2017attention}, by allowing the tokens to interact with each other (“self”) and find out who they should pay more attention to (“attention”), and the addition of positional information of tokens further increases the use of sequential order information. Excitingly, the Vison Transformer (ViT) enables the application of transformers to 2D images \cite{dosovitskiy2020vit}. Inspired by these studies, we propose a transformer layer to interpret our graph-structured data. While the SA mechanism has been extensively used in the context of natural language processing, we extended the framework for WSI data. Briefly, the standard \textbf{qkv} self-attention \cite{vaswani2017attention} is a mechanism to find the words of importance for a given query word in a sentence, and it receives as input a 1D sequence of token embeddings. For the graph, the feature nodes are treated as tokens in a sequence and the adjacency matrix is used to denote the positional information. Given that $\boldsymbol{\rm x} \in \mathbb{R}^{N \times D}$ is the sequence of patches (feature nodes) in the graph, where $N$ is the number of patches and $D$ is the embedding dimension of each patch, we compute \textbf{q}(query), \textbf{k}(key) and \textbf{v}(value) (Eq.\ref{eq:6a}). The attention weights $A_{ij}$ are based on the pairwise similarity between two patches of the sequence and their respective query $\boldsymbol{\rm q}^{i}$ and key $\boldsymbol{\rm k}^{j}$ in Eq.\ref{eq:6b}. Multihead Self-Attention (MSA) is a mechanism that involves combining the knowledge explored by $k$ number of SA operations, called \textquoteleft heads\textquoteright. It projects concatenated outputs of SA in Eq.\ref{eq:6c}. $D_h$ (Eq.\ref{eq:6a}) is typically set to $D/k$ to facilitate computation and maintain the number of parameters constant when changing $k$.
\begin{subequations}\label{eq:6}
    \begin{align}
    \label{eq:6a}
    &  [\boldsymbol{\rm q}, \boldsymbol{\rm k}, \boldsymbol{\rm v}]=\boldsymbol{\rm x} \boldsymbol{\rm U}_{qkv}, \qquad \boldsymbol{\rm U}_{qkv} \in  \mathbb{R}^{D \times 3D_h}\\
    \label{eq:6b}
    & A= {\rm softmax}(\boldsymbol{\rm q}\boldsymbol{\rm k}^{T}/\sqrt{D_h}), \qquad A \in \mathbb{R}^{N \times N}\\
    \label{eq:6c}
    & {\rm SA(\boldsymbol{\rm x})}= A \boldsymbol{\rm v}, \\
    \label{eq:6d}
    \begin{split}
    & {\rm MSA(\boldsymbol{\rm x})}= [{\rm SA_1 \boldsymbol{(\rm x)};SA_2 \boldsymbol{(\rm x)}; \ldots SA_k \boldsymbol{(\rm x)}}] \boldsymbol{\rm U}_{msa}, \text{and} \\ 
    & \qquad \boldsymbol{\rm U}_{msa} \in \mathbb{R}^{k \cdot D_h \times D}.
    \end{split}
    \end{align}
\end{subequations}
The goal of the transformer layer is to learn the mapping: $\mathbb{H} \rightarrow \mathbb{T}$, where $\mathbb{H}$ is the graph space, and $\mathbb{T}$ is the transformer space. We define the mapping of $\mathbb{H} \rightarrow \mathbb{T}$ as:
    \begin{subequations}\label{eq:3}
    \begin{align}
    \label{eq:3a}
    & t_{0} = [x_{\rm class}; h^{(1)}; h^{(2)}; \: \ldots ; \:
        h^{(N)}], \quad \quad h^{(i)} \in H \\
    \label{eq:3b}
    & t^{'}_{l} = {\rm MSA}({\rm LN}(t_{l-1})) + t_{l-1}, \quad \quad l=1 \ldots L \\
    \label{eq:3c}
    & t_{l} = {\rm MLP}({\rm LN}(t^{'}_{l})) + t^{'}_{l}, \quad \quad l=1 \ldots L 
    \end{align}
    \end{subequations}
where MSA is the Multiheaded Self-Attention (Eq.\ref{eq:6}), MLP is a Multilayer Perceptron, and LN denotes Layer Norm. L is the number of MSA blocks \cite{dosovitskiy2020vit}. Each block consists of an MSA layer (Eq.\ref{eq:3b}) and an MLP block (Eq.\ref{eq:3c}). In order to learn the mapping $\mathbb{T} \rightarrow \mathbb{Y}$ from transformer space $\mathbb{T}$  to label space $\mathbb{Y}$, we prepared a learnable embedding ($t_0^{(0)}=x_{\rm class}$) to the feature nodes (Eq.\ref{eq:3a}), whose state at the output of the transformer layer ($z_{L}^0$) serves as mapping of $\mathbb{T} \rightarrow \mathbb{Y}$:
    \begin{equation}
        y= LN(z_{L}^{(0)}).
    \end{equation}

Recently, position embeddings were added to the patch embeddings to retain positional information  \cite{dosovitskiy2020vit}. Position embedding explores absolute position encoding (e.g., sinusoidal or learnable absolute encoding) as well as conditional position encoding. However, the learnable absolute encoding is commonly used in problems with fixed length sequences and does not meet the requirement for WSI analysis. This is because the number of patches that can be extracted can vary among different WSIs. To address this, Islam and colleagues showed that the addition of zero padding can provide an absolute position information for convolution \cite{islam2020position}. In our work, we do not leverage position embeddings in the same fashion as ViTs or via inclusion of zero padding. Rather, we used position embedding in the form of an adjacency matrix for the graph convolutional layers. The adjacency matrix of the WSI graph accounts for the spatial information and inter-connectivity of the nodes, which is preserved when performing graph convolutions. To reduce the number of inputs to the transformer layer, a pooling layer was added between the graph convolution and transformer layer (see below). The positional information between the pooled adjacency matrix and the original patch adjacency matrix is preserved by a dense learned assignment used in pooling the features. With this framework, we were able to avoid the need of adding an additional encoder in the transformer and instead preserved positional information in the GC layers of GTP, thus reducing the complexity of our model.

The softmax function is typically used as a row-by-row normalization function in transformers for computer vision tasks \cite{zhu2021deformable, chen2021transunet}. The standard self-attention mechanism requires the calculation of similarity scores between each pair of nodes, resulting in both memory and time complexity that is quadratic in the number of nodes ($O(n^2)$). Since the WSI graph contains several nodes (in thousands), it may lead to out-of-memory issues during model training and thus applying the transformer layer directly to the convolved graphs is not trivial. We therefore added a mincut pooling layer \cite{bianchi2020spectral} between the GC and transformer layers that reduced the number of inputs from thousands to hundreds of nodes. Alternatively, mean pooling can be used to reduce the number of nodes but it may lose information. As such, the mean pooling operation does not contain learnable parameters like what we have with the case of min-cut pooling. Min-cut pooling could preserve the local information of neighboring nodes and reduce the number of nodes at the same time. The idea behind min-cut pooling is to take a continuous relaxation of the min-cut problem and implement it as a pooling layer with a custom loss function. By minimizing the custom loss, the pooling layer learns to find min-cut clusters on any given graph and aggregates the clusters to reduce the graph size. At the same time, because the pooling layer can be used as part of the entire GTP, the classification loss of GTP that is being minimized during training will influence the min-cut layer as well. In so doing, our GTP graph-transformer was able to accommodate several patches as input, underscoring the novelty of our approach and its application to WSI data.

\begin{figure}[hbt!]
\centerline{\includegraphics[width=\linewidth]{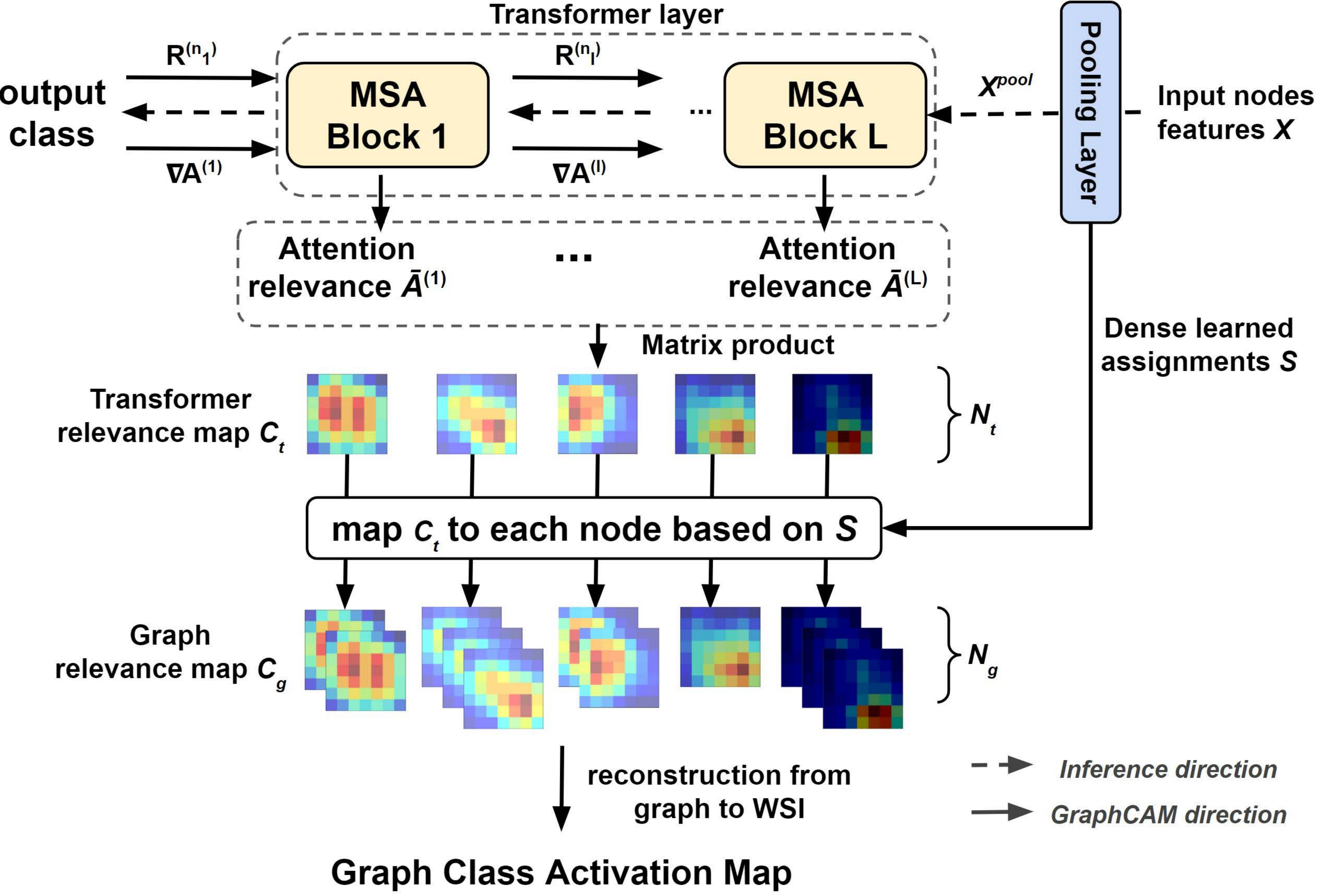}}
\caption{\textbf{Schematic of the GraphCAM.} Gradients and relevance are propagated through the network and integrated with an attention map to produce the transformer relevancy maps. Transformer relevancy maps are then mapped to graph class activation maps via reverse pooling.}
\label{fig:graphCAM}
\vspace{-3mm}
\end{figure}

\subsection{Class activation mapping}
To understand how GTP processes WSI data and identifies regions that are highly associated with the class label, we proposed a novel class activation mapping technique on graphs (Fig.~\ref{fig:graphCAM}). In what follows, we use the term GraphCAM to refer to this technique. Our technique was inspired by the recent work by Chefer and colleagues \cite{chefer2021transformer}, who used the deep Taylor decomposition principle to assign local relevance scores and propagated them through the layers by maintaining the total relevancy across layers. In a similar fashion, our method computes the class activation map from the output class to the input graph space, and reconstructs the final class activation map for the WSI from its graph representation.

Let $A^{(l)}$ represent the attention map of the MSA block $l$ in Eq.\ref{eq:6b}. Following the propagation procedure of relevance and gradients by Chefer and colleagues \cite{chefer2021transformer}, 
GraphCAM computes the gradient $\nabla A^{(l)}$ and layer relevance $R^{(n_l)}$ with respect to a target class for each attention map $A^{(l)}$, where $n_l$ is the layer that corresponds to the softmax operation in Eq.\ref{eq:6b} of block $l$. The transformer relevance map $C_t$ is then defined as a weighted attention relevance:
\begin{subequations}\label{eq:5}
\begin{align}
\label{eq:5a}
& C_{t} = \prod_{l=1}^{L} \bar{A}^{(l)} \\
\label{eq:5b}
& \bar{A}^{(l)} = \mathbb{E}_{h}(\nabla A^{(l)} \odot R^{(n_l)}) + I 
\end{align}
\end{subequations}
where $\odot$ is the Hadamard product, $\mathbb{E}_{h}$ is the mean across the \textquoteleft heads\textquoteright~dimension, and $I$ is the identity matrix to avoid self inhibition for each node.

The pooled node features by the mincut pooling layer are computed as $X^{pool}=S^{T}X$, where $S \in \mathbb{R}^{N_g \times N_t}$ is the dense learned assignment, and $N_t$ and $N_g$ are the number of nodes before and after the pooling layer. To yield the graph relevance map $C_g$ from transformer relevance map $C_t$, our GraphCAM performs mapping $C_t$ to each node in the graph based on the dense learned assignments as $C_t \xrightarrow{S} C_g$. Finally, GraphCAM reconstructs the final class activation map on the WSI using the adjacency matrix of the graph and coordinates of the patches.

\subsection{Data and code availability}
All the WSIs and corresponding clinical data can be downloaded freely from CPTAC, TCGA and NLST websites. Python scripts and manuals are made available on GitHub (\url{https://github.com/vkola-lab/tmi2022}).

\begin{figure*}[t]
    \centering
    \begin{subfigure}[b]{0.45\textwidth}
        \centering
        \includegraphics[width=\textwidth]{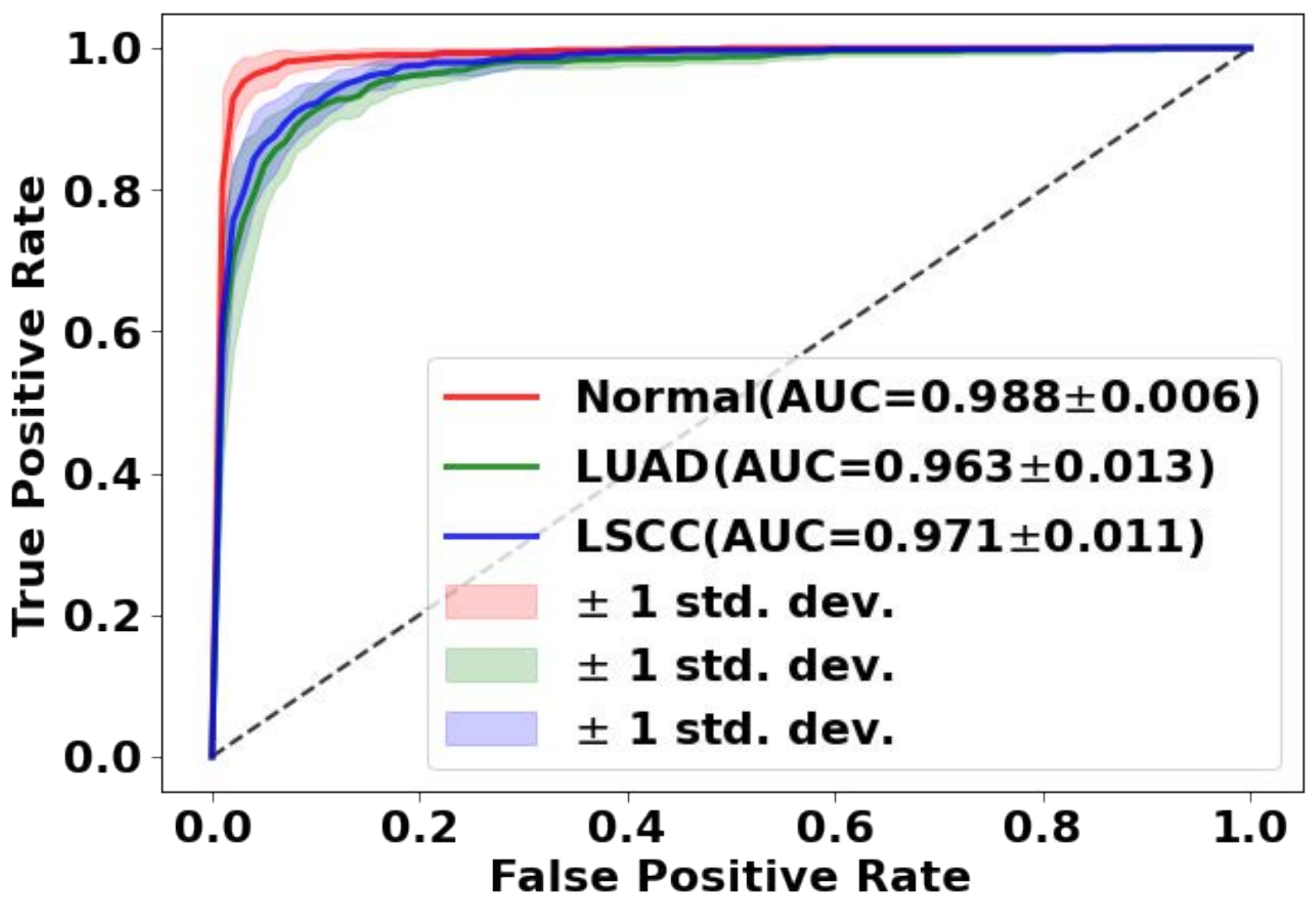}
        \caption[]%
        {{\small CPTAC ROC curves}}    
        \label{}
    \end{subfigure}
    \qquad
    \begin{subfigure}[b]{0.45\textwidth}  
        \centering 
        \includegraphics[width=\textwidth]{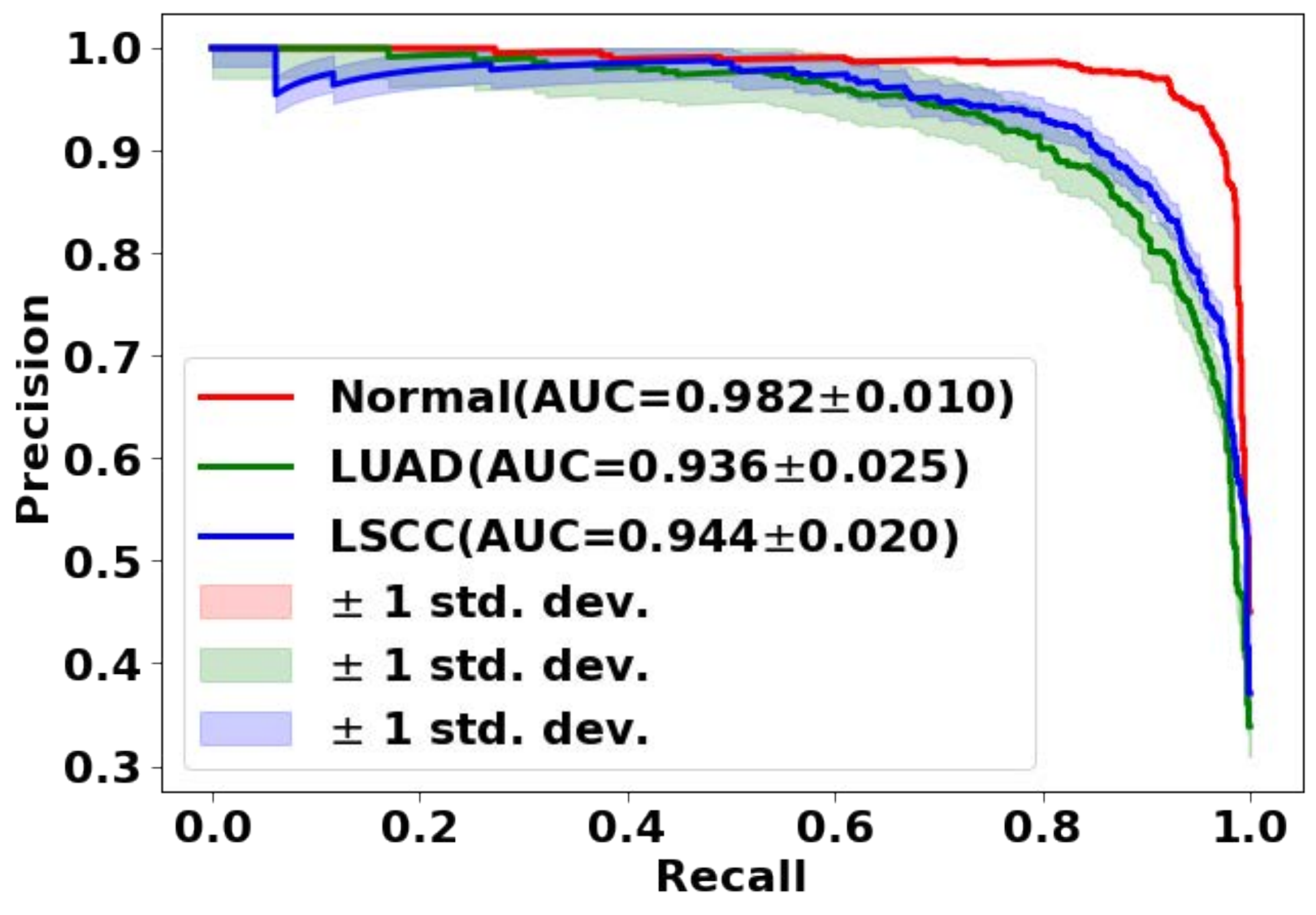}
        \caption[]%
        {{\small CPTAC PR curves}}    
        \label{}
    \end{subfigure}
    \begin{subfigure}[b]{0.45\textwidth}   
        \centering 
        \includegraphics[width=\textwidth]{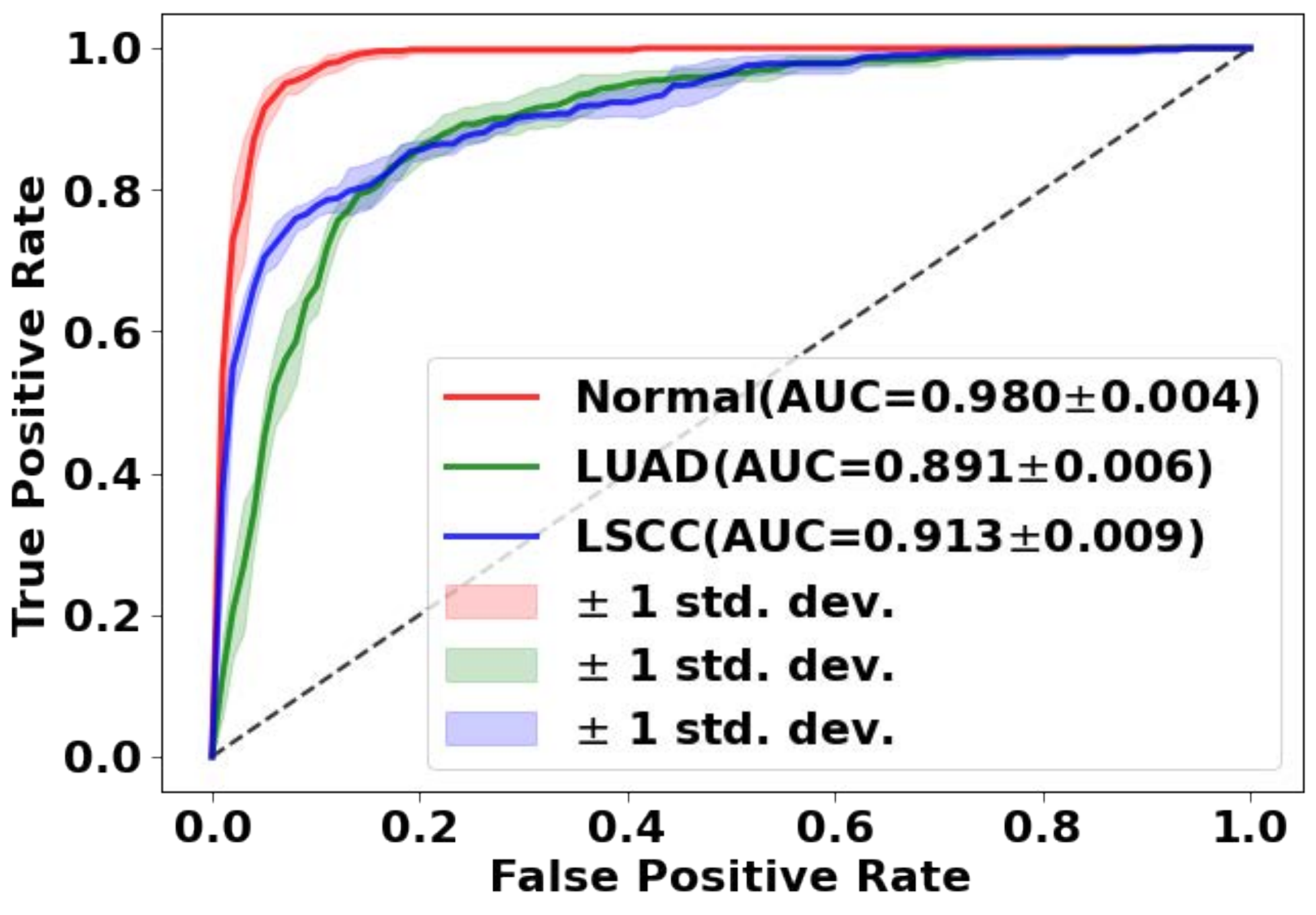}
        \caption[]%
        {{\small TCGA ROC curves}}    
        \label{}
    \end{subfigure}
    \qquad
    \begin{subfigure}[b]{0.45\textwidth}   
        \centering 
        \includegraphics[width=\textwidth]{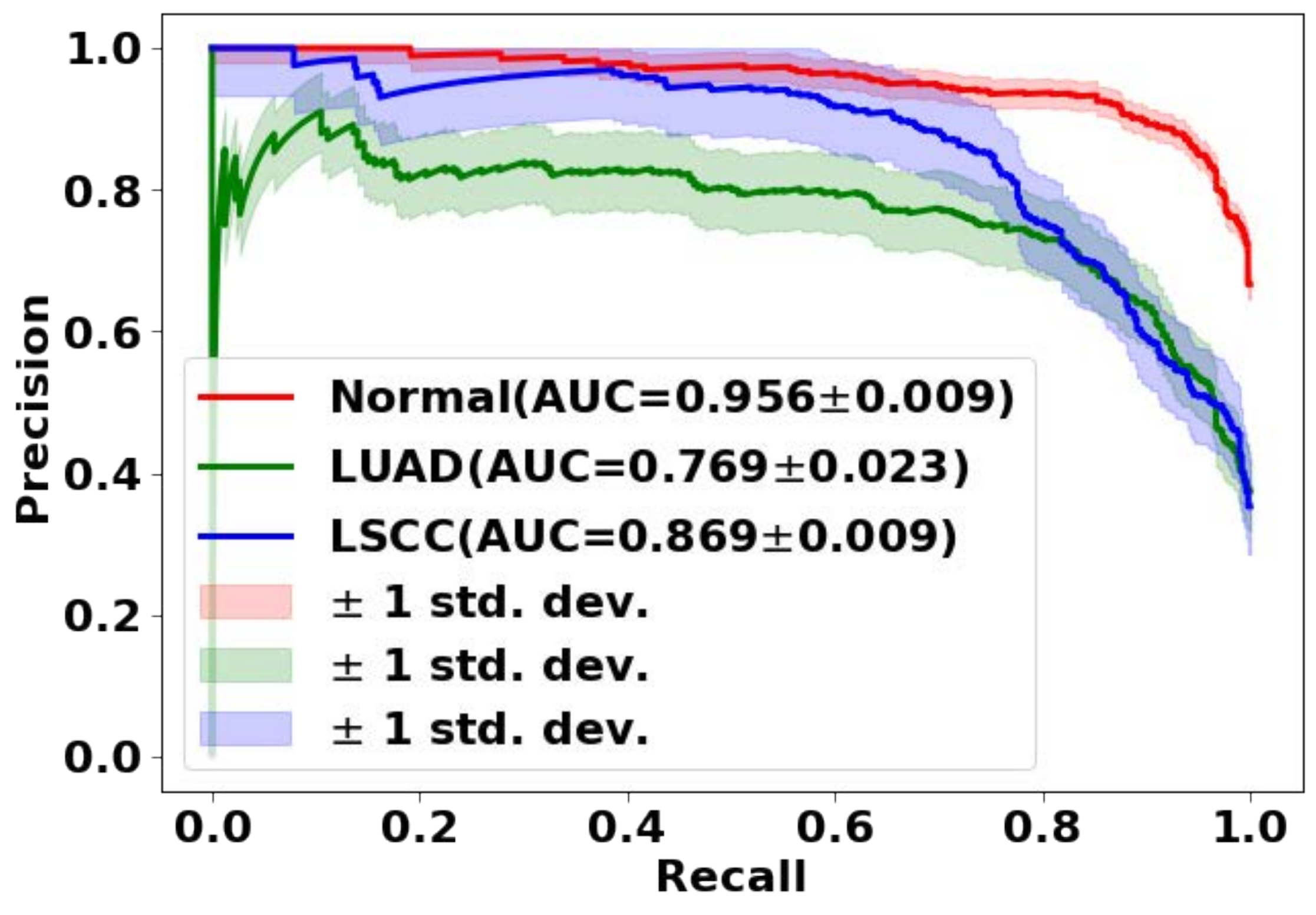}
        \caption[]%
        {{\small TCGA PR curves}}    
        \label{}
    \end{subfigure}
    \caption[ The average and standard deviation of critical parameters ]
    {\small \textbf{Model performance on the CPTAC and TCGA datasets}. Mean ROC and PR curves along with standard deviations for the classification tasks (normal vs. tumor; LUAD vs. others; LSCC vs. others) are shown.} 
    \label{fig:roc_pr_3_label}
\end{figure*}

\begin{table*}[]
\caption{\textbf{Performance metrics for the 3-label (Normal vs. LUAD vs. LSCC) classification task.} Mean performance metrics are reported along with the corresponding values of standard deviation in parentheses.}
\label{table:metrics}
\centering
\setlength\tabcolsep{2pt} 
\subfloat[Precision, Recall/Sensitivity, and Specificity (Percentage (\%) values are reported).]{
\begin{tabular}{c|c|ccc|ccc|ccc}
\hline
\multirow{2}{*}{Method} & \multirow{2}{*}{Data} & \multicolumn{3}{c|}{Precision} & \multicolumn{3}{c|}{Recall/Sensitivity}  & \multicolumn{3}{c}{Specificity}  \\
 & & Normal     & LUAD        & LSCC        & Normal     & LUAD        & LSCC        & Normal     & LUAD        & LSCC  \\ \hline
\multirow{2}{*}{\begin{tabular}[c]{@{}c@{}}TransMIL\\ \cite{transmil} \end{tabular}}  & CPTAC  & $89.7(1.8)$ & $81.0(3.1)$ & $87.1(1.7)$ & $90.4(1.9)$ & $81.2(2.0)$ & $85.9(3.7)$ & $94.4(1.1)$ & $90.8(1.9)$ & $93.7(1.2)$  \\
\multicolumn{1}{c|}{}                          & TCGA  & $76.6(5.1)$ & $64.3(4.1)$ & $80.4(1.7)$ & $87.3(3.6)$ & $75.8(5.3)$ & $55.6(7.7)$ & $90.6(2.9)$ & $73.0(5.6)$ & $92.4(1.9)$ \\ \hline

\multirow{2}{*}{\begin{tabular}[c]{@{}c@{}}AttPool\\ \cite{huang2019attpool} \end{tabular}}  & CPTAC  & $88.4(2.7)$ & $77.1(2.7)$ & $80.6(3.3)$ & $85.9(3.3)$ & $78.0(4.7)$ & $81.6(1.8)$ & $94.0(1.5)$ & $88.9(2.1)$ & $90.1(2.2)$  \\
\multicolumn{1}{c|}{}                          & TCGA  & $89.1(4.1)$ & $69.9(3.9)$ & $81.4(3.0)$ & $87.8(2.7)$ & $79.4(2.4)$ & $71.3(4.0)$ & $\mathbf{94.9(2.0)}$ & $82.5(2.9)$ & $91.4(1.7)$ \\ \hline

\multirow{2}{*}{\begin{tabular}[c]{@{}c@{}}GTP$^{\star}$\\ (only GCN)\end{tabular}}  & CPTAC  & $82.9(6.5)$ & $81.6(6.5)$ & $86.5(4.7)$ & $93.6(5.0)$ & $74.4(6.7)$ & $80.3(4.8)$ & $89.3(5.3)$ & $91.6(3.9)$ & $93.6(2.7)$  \\
\multicolumn{1}{c|}{}                          & TCGA  & $72.5(9.0)$ & $69.1(3.2)$ & $82.0(9.8)$ & $90.7(8.3)$ & $57.7(9.7)$ & $69.9(9.5)$ & $82.4(9.9)$ & $85.7(8.4)$ & $90.7(6.1)$ \\ \hline

\multicolumn{1}{c|}{\multirow{2}{*}{\textbf{GTP}}}  & CPTAC  & $\mathbf{93.2(3.0)}$ & $\mathbf{88.4(3.9)}$ & $\mathbf{87.8(3.0)}$ & $\mathbf{95.9(2.2)}$ & $\mathbf{83.9(4.5)}$ & $\mathbf{89.2(4.0)}$ & $\mathbf{96.2(1.7)}$ & $\mathbf{94.7(1.9)}$ & $\mathbf{93.8(1.7)}$  \\
\multicolumn{1}{c|}{}                      & TCGA  & $\mathbf{89.2(2.8)}$ & $\mathbf{74.4(2.7)}$ & $\mathbf{84.4(0.7)}$ & $\mathbf{92.6(2.7)}$ & $\mathbf{79.8(1.9)}$ & $\mathbf{75.2(1.6)}$ & $94.7(1.6)$ & $\mathbf{86.0(2.3)}$ & $\mathbf{92.7(0.4)}$ \\ \hline
\end{tabular}}

\vspace{0.3cm}

\centering
\setlength\tabcolsep{5pt} 
    \begin{subtable}{.5\linewidth}
      \centering
        \caption{Accuracy and AUC (Percentage (\%) values are reported).}
        \begin{tabular}{c|c|c|c}
        \hline
        \multicolumn{1}{c|}{Method} & \multicolumn{1}{c|}{Data} & \multicolumn{1}{c|}{Accuracy} & \multicolumn{1}{c}{AUC} \\ \hline
        \multirow{2}{*}{TransMIL\cite{transmil}}  & CPTAC                        & $85.9(0.7)$ & $96.1(0.3)$   \\
                                     & TCGA                          & $71.6(2.3)$ & $88.0(0.7)$   \\ \hline
        \multirow{2}{*}{AttPool\cite{huang2019attpool}}   & CPTAC                         & $81.9(2.1)$ & $92.5(1.6)$   \\
                                     & TCGA                          & $79.3(2.3)$ & $91.3(1.1)$   \\ \hline
        \multirow{2}{*}{\begin{tabular}[c]{@{}c@{}}GTP$^{\star}$\\ (only GCN)\end{tabular}}   & CPTAC    & $83.0(2.7)$ & $95.2(1.2)$   \\
                                   & TCGA                          & $72.4(4.9)$ & $86.6(3.8)$   \\ \hline 
        \multirow{2}{*}{\textbf{GTP}}       & CPTAC                               & $\mathbf{91.2(2.5)}$ & $\mathbf{97.7(0.9)}$   \\
                                   & TCGA                          & $\mathbf{82.3(1.0)}$ & $\mathbf{92.8(0.3)}$   \\ \hline
        \end{tabular}
    \end{subtable}%
    \begin{subtable}{.5\linewidth}
      \centering
        \caption{DeLong's algorithm for comparing the AUC values between GTP and other methods. $\text{log}_{10}(0.05)=-1.301$.}
        \begin{tabular}{c|c|c}
        \hline
        \multicolumn{1}{c|}{Method} & \multicolumn{1}{c|}{Data} & \multicolumn{1}{c}{$\text{log}_{10}$(p-value)}  \\ \hline
        \multirow{2}{*}{TransMIL\cite{transmil}}  & CPTAC                          & $-1.578(0.853)$     \\
                                     & TCGA                          & $-5.627(2.263)$    \\ \hline
        \multirow{2}{*}{AttPool\cite{huang2019attpool}}   & CPTAC                          & $-2.305(1.250)$    \\
                                     & TCGA                          & $-2.068(1.339)$    \\ \hline
        \multirow{2}{*}{\begin{tabular}[c]{@{}c@{}}GTP$^{\star}$\\ (only GCN)\end{tabular}}   & CPTAC  & $-1.759(1.129)$   \\
                                   & TCGA                          & $-5.373(3.146)$     \\ \hline 
        \end{tabular}
    \end{subtable}
\end{table*}

\begin{figure*}[htp]
\centering
\setlength\tabcolsep{2pt} 
\begin{tabular}{|c|c|c|c|c|}
\hline
\bf{Whole Slide Image} & \bf{GraphCAM} & \bf{Annotation} & \bf{Binarized GraphCAM} & \bf{IoU} \\ \hline
\includegraphics[width=.36\columnwidth, keepaspectratio=true]{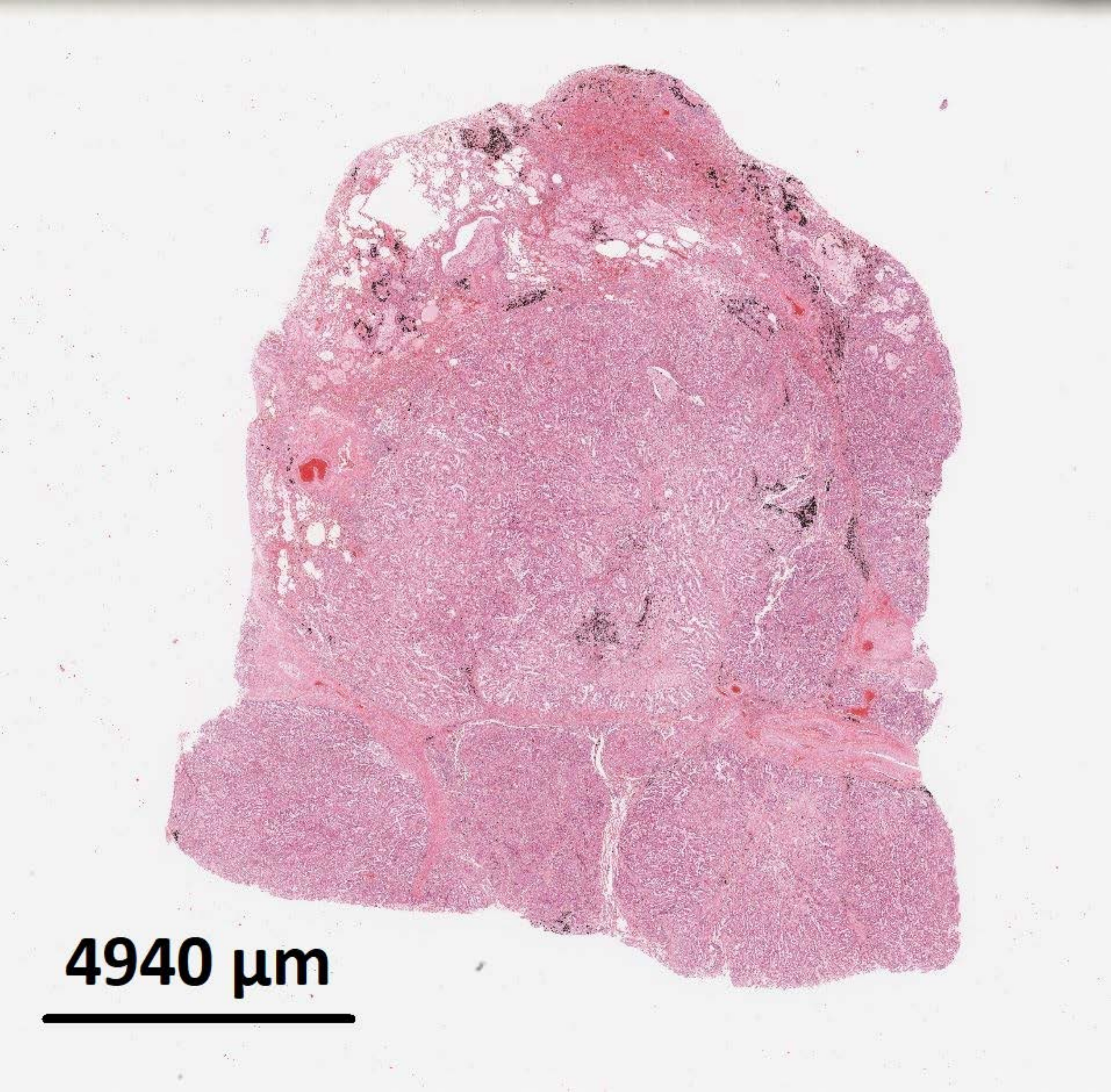} &
\includegraphics[width=.36\columnwidth, keepaspectratio=true]{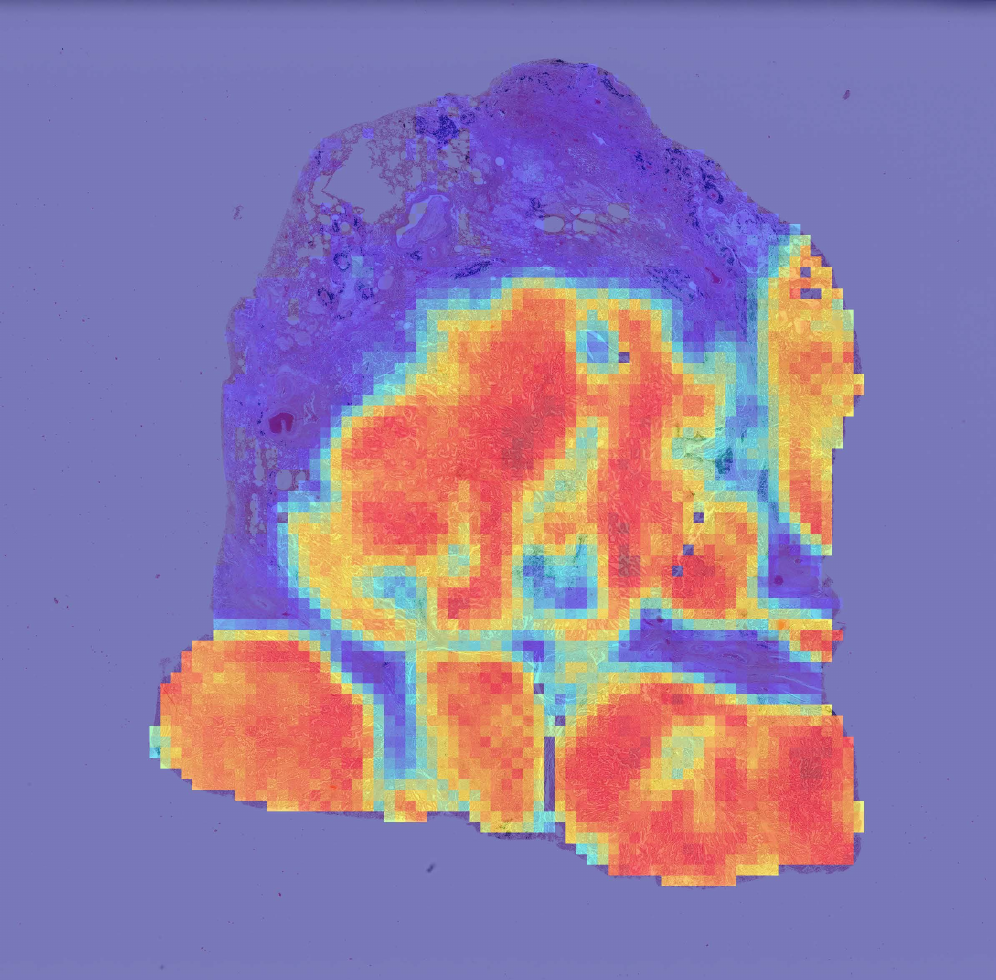} &
\includegraphics[width=.36\columnwidth, keepaspectratio=true]{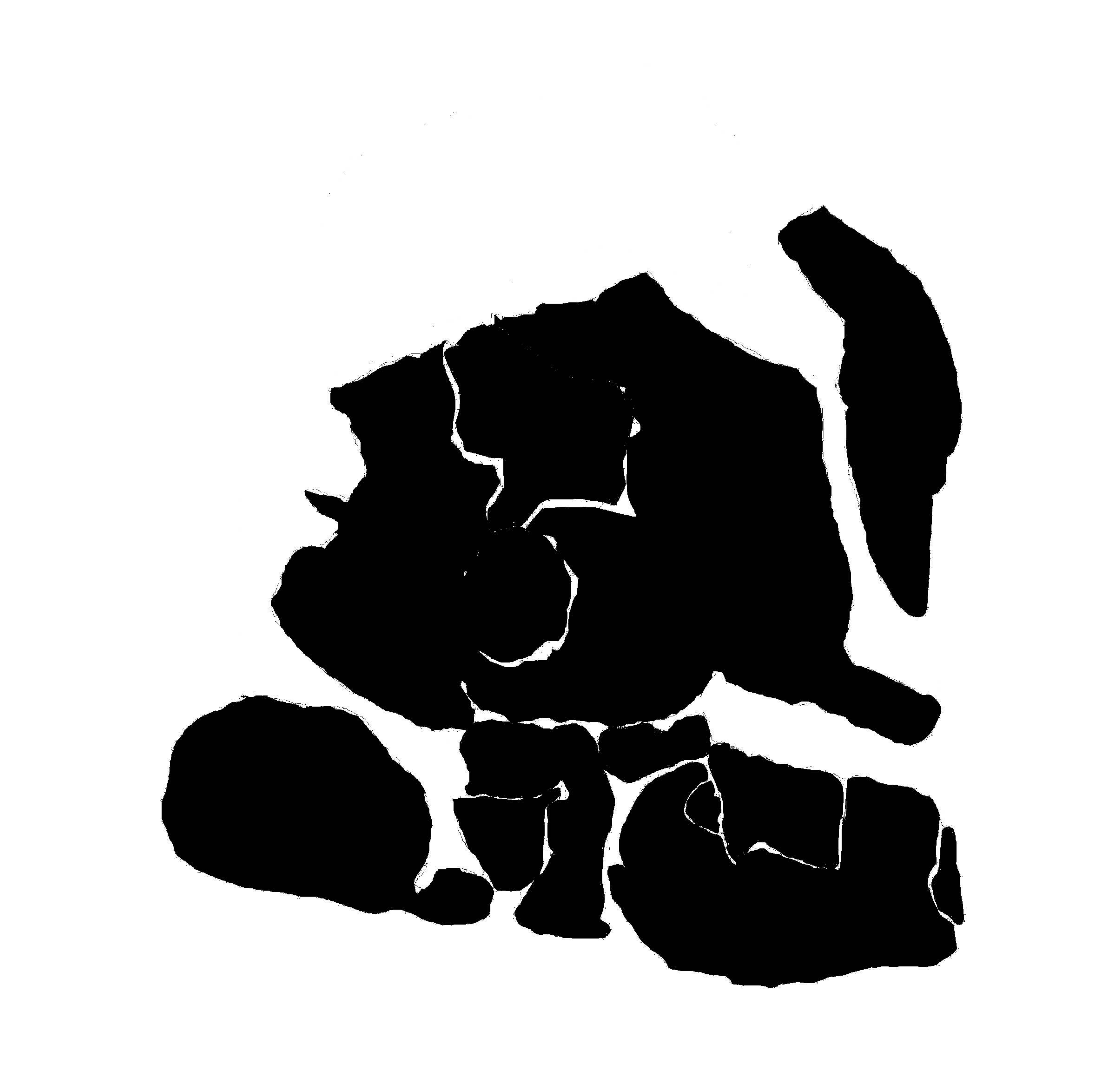} &
\includegraphics[width=.36\columnwidth, keepaspectratio=true]{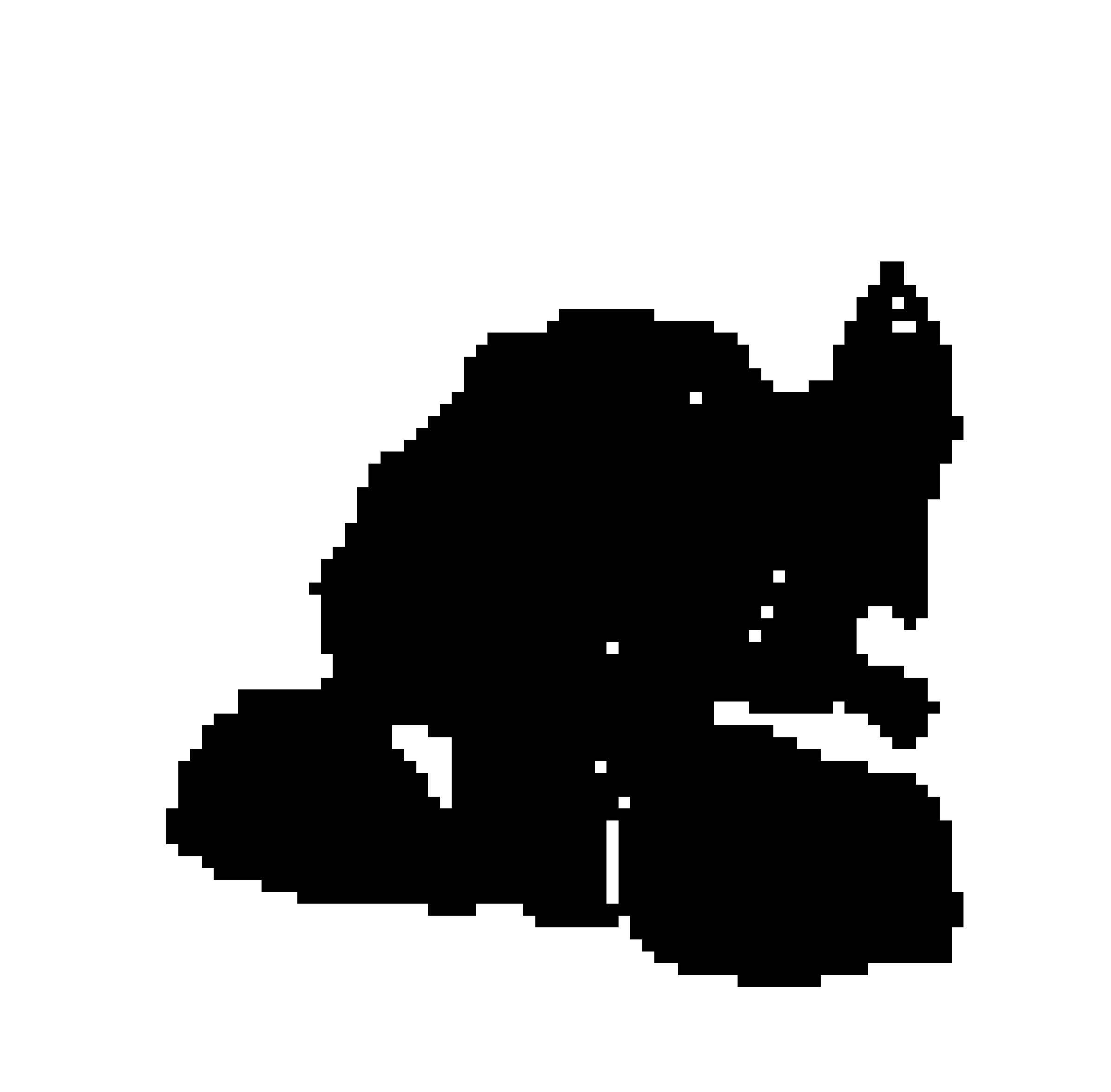} &
\includegraphics[width=.50\columnwidth, keepaspectratio=True]{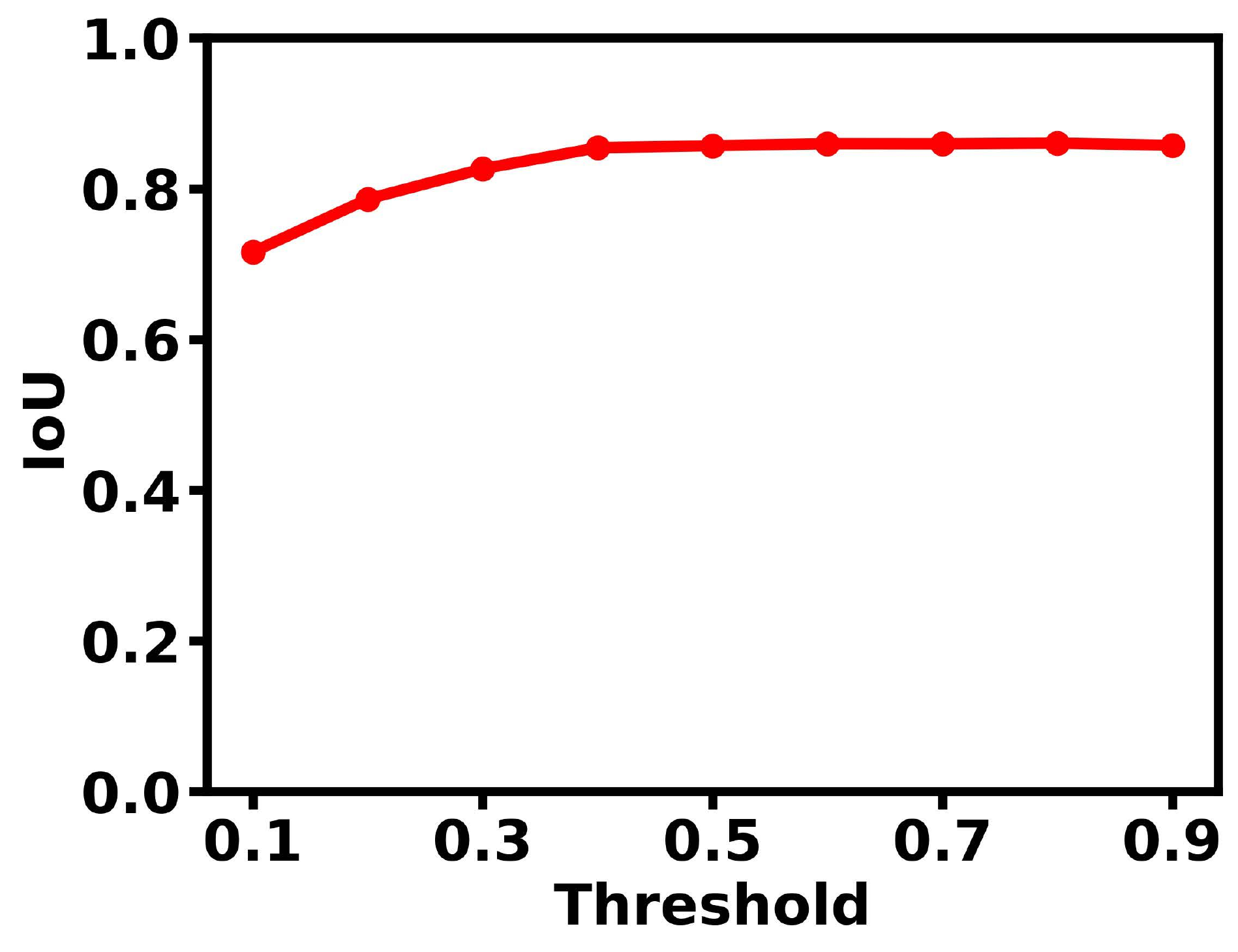}\\
\hline
\raisebox{.2\height}{\includegraphics[width=.36\columnwidth, keepaspectratio=true]{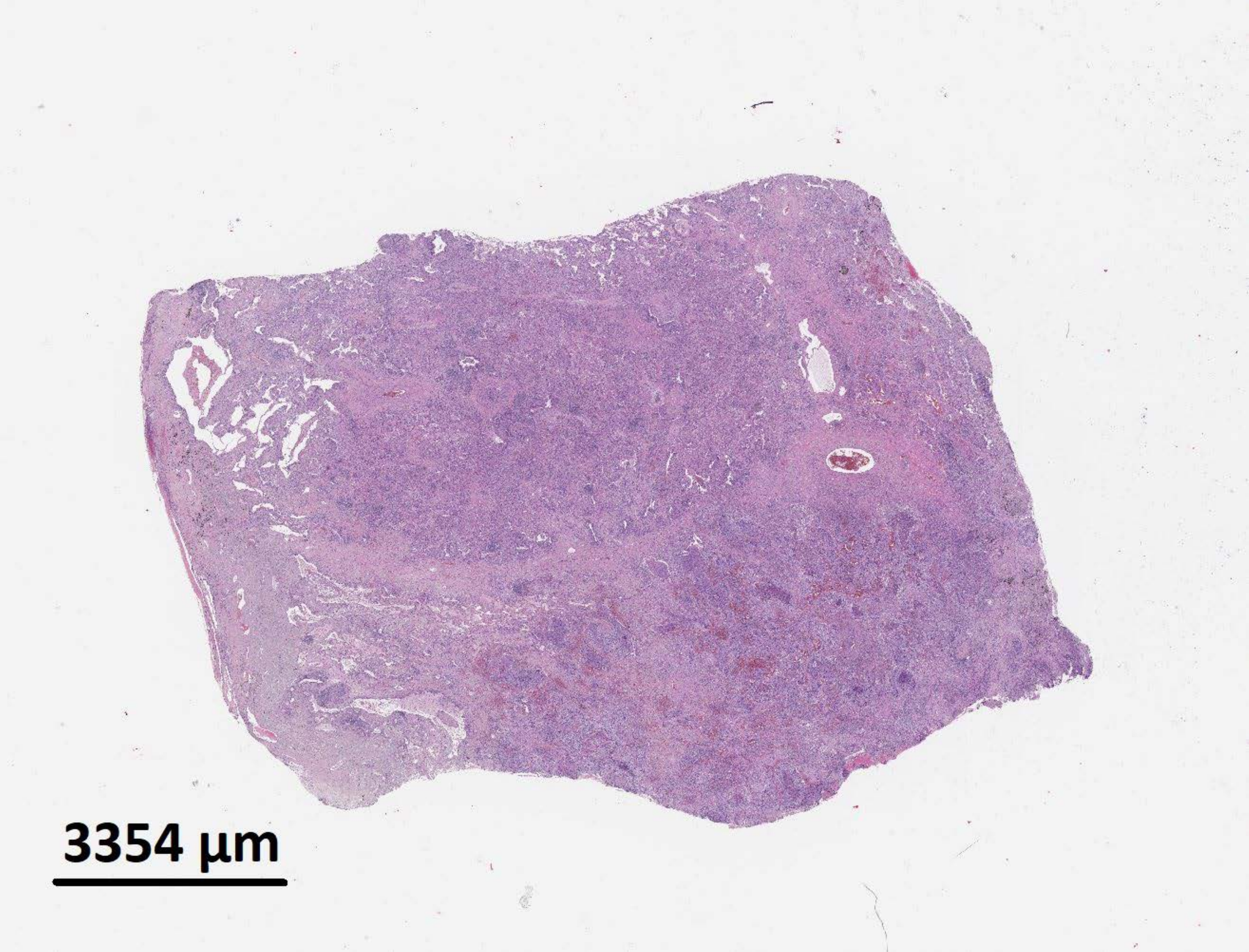}} &
\raisebox{.2\height}{\includegraphics[width=.36\columnwidth,keepaspectratio=true]{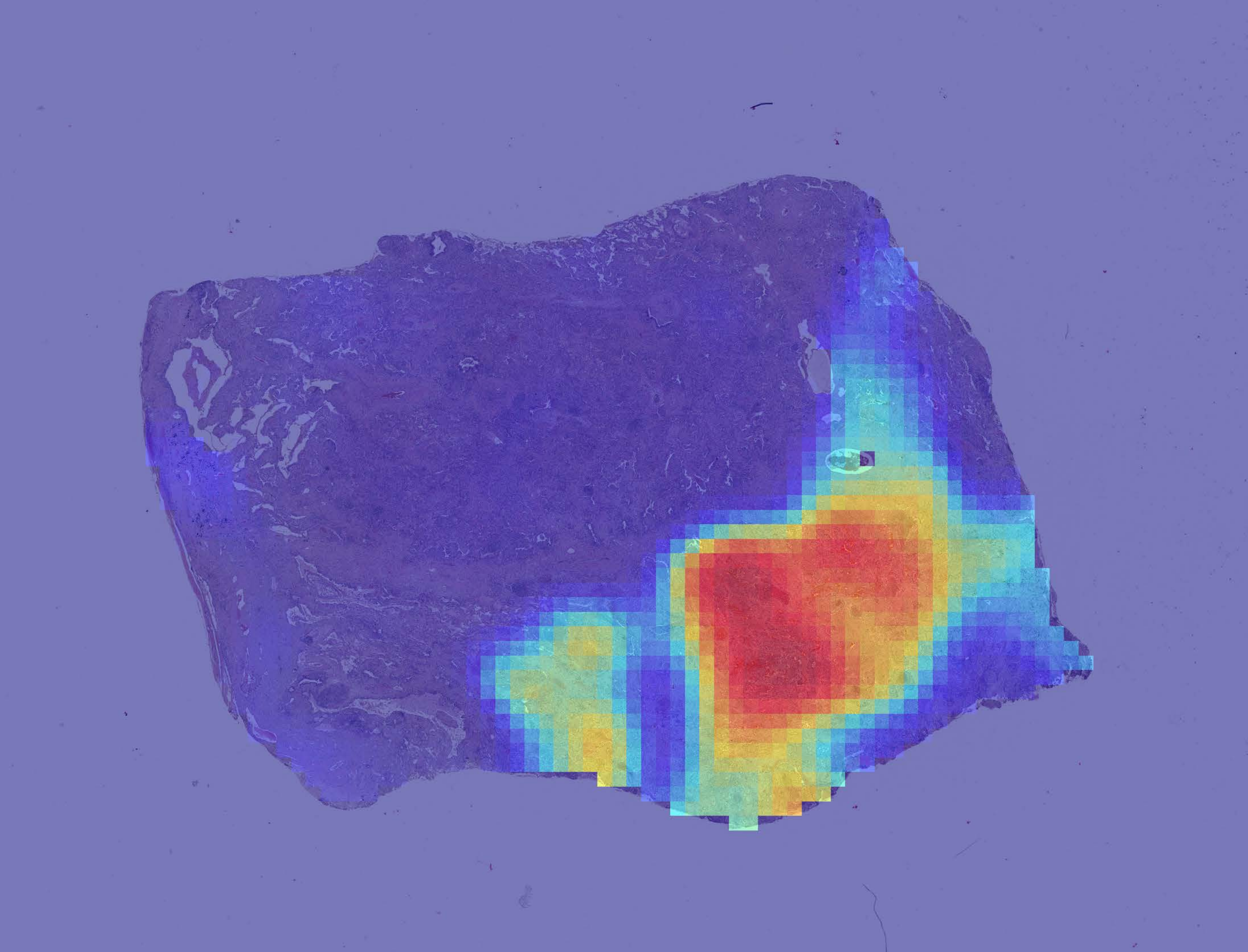}} &
\raisebox{.2\height}{\includegraphics[width=.36\columnwidth, keepaspectratio=true]{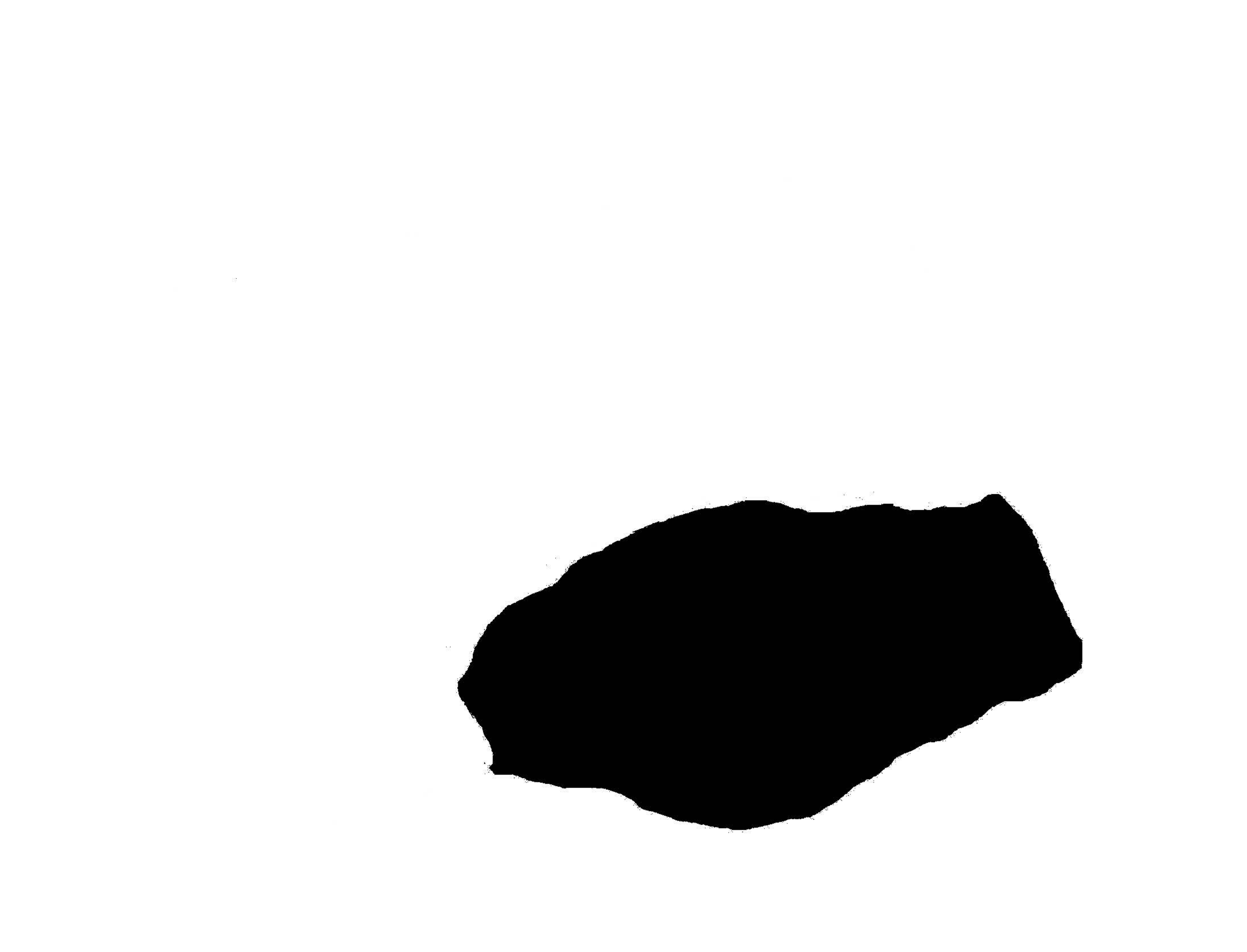}} &
\raisebox{.2\height}{\includegraphics[width=.36\columnwidth, keepaspectratio=true]{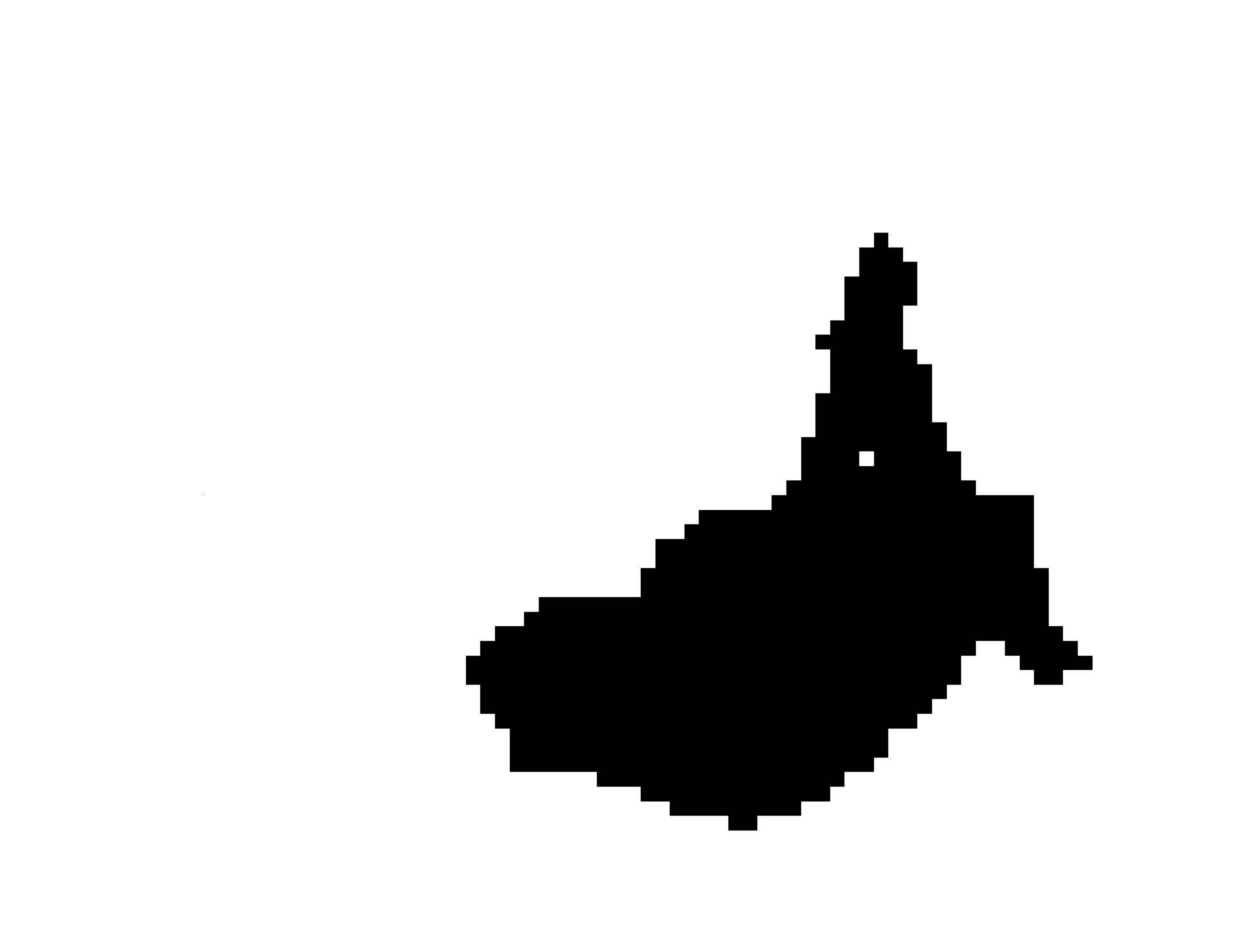}} &
\includegraphics[width=.50\columnwidth, keepaspectratio=True]{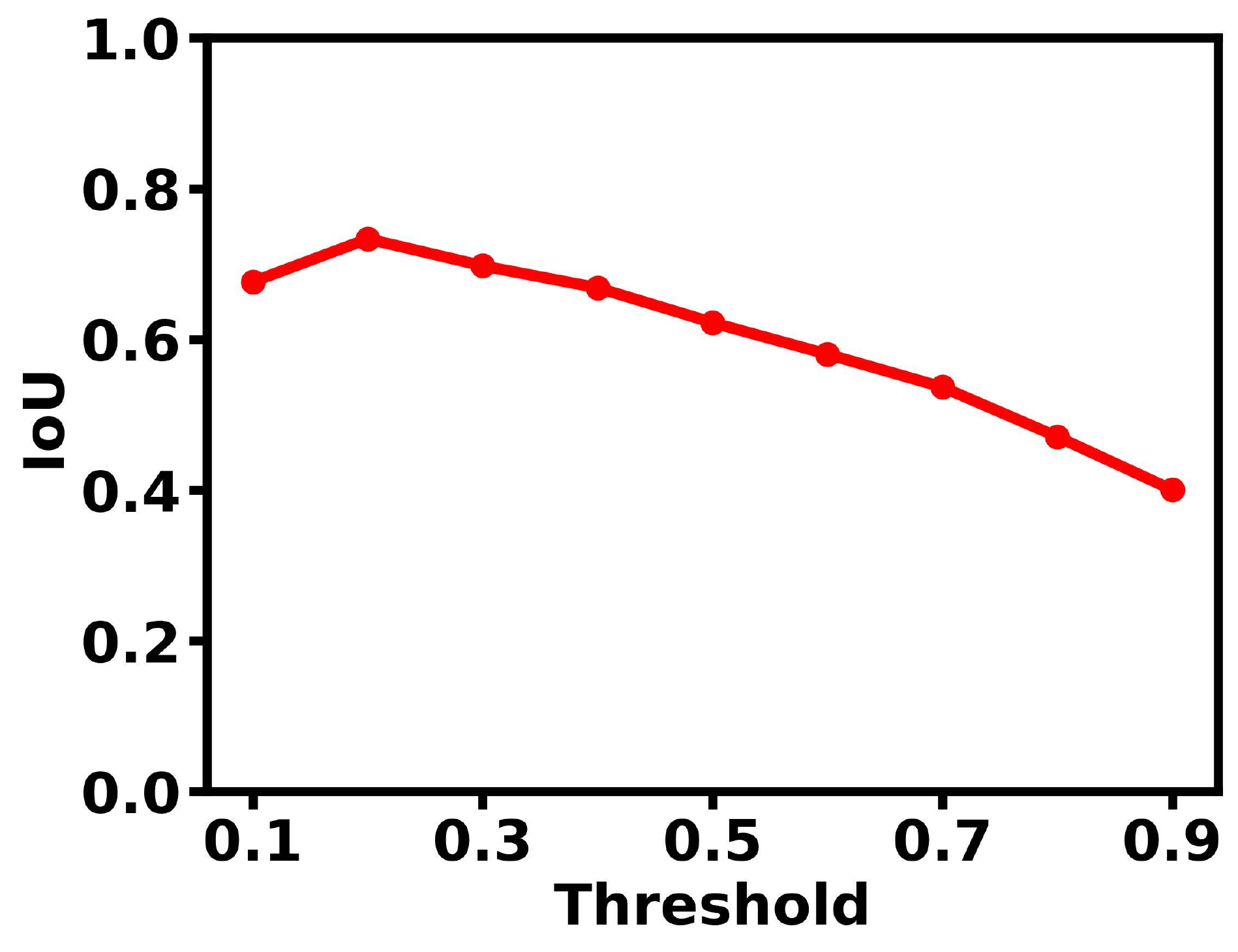}\\
\hline
& \multicolumn{1}{r|}{\includegraphics[width=.06\textheight, height=0.02\textheight]{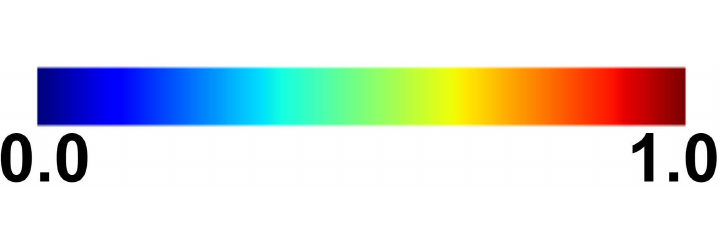}} & & & \\
\hline
\end{tabular}
\caption{\textbf{GraphCAMs and their comparison with the expert annotations.} For each WSI, we generated GraphCAMs and compared them with annotations from the pathologist. The first column contains the original WSIs, the second and third columns contain GraphCAMs and pathologist's annotations, respectively and the fourth column contains the binarized GraphCAMs based on the threshold from the Intersection of Union (IoU) plot in the last column. The first row shows an LUAD case and the second row denotes an LSCC case.}
\label{fig:cam}
\end{figure*}

\begin{figure*}[htb]
    \setlength\tabcolsep{2pt} 
    \centering
    \begin{tabular}{|c | c c c c c|}
      \hline
      \bf{Whole Slide Image}  & \multicolumn{5}{c|}{\bf{GraphCAM from cross-validation models} } \\
      \includegraphics[width=.155\textwidth,height=.12\textheight]{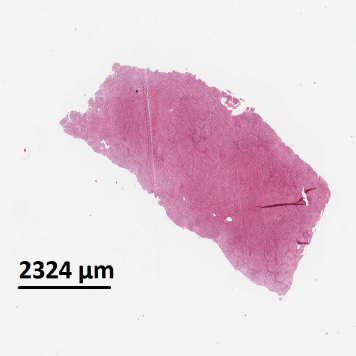} &
      \includegraphics[width=.155\textwidth,height=.12\textheight]{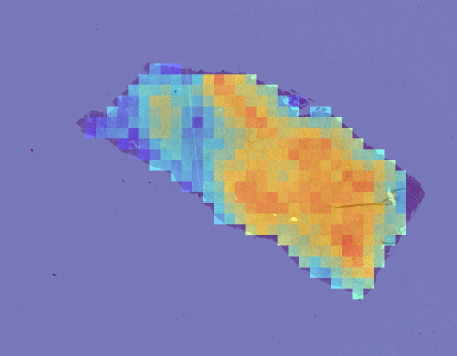} &
      \includegraphics[width=.155\textwidth,height=.12\textheight]{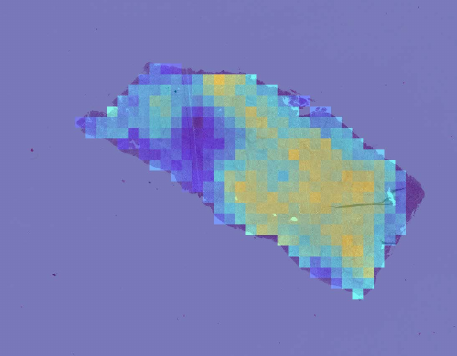} &
      \includegraphics[width=.155\textwidth,height=.12\textheight]{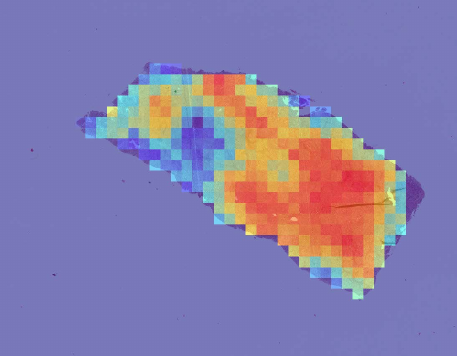} &
      \includegraphics[width=.155\textwidth,height=.12\textheight]{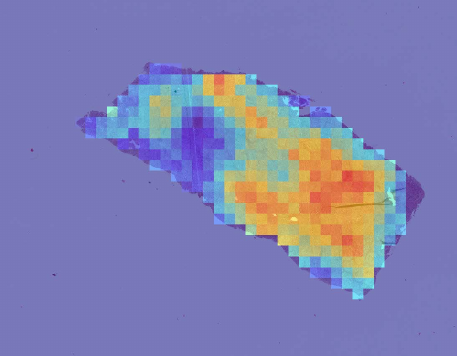} &
      \includegraphics[width=.155\textwidth,height=.12\textheight]{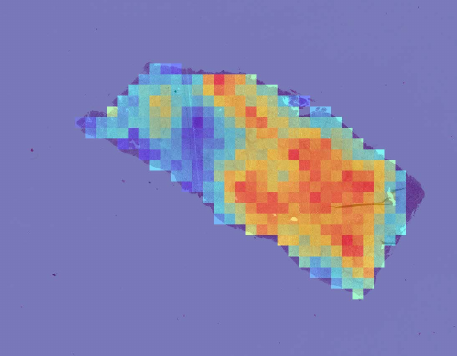} \\
      LUAD & Model 1 (p=0.77) & Model 2 (p=0.62) & Model 3 (p=0.87) & Model 4 (p=0.81) & Model 5 (p=0.88) \\
      \includegraphics[width=.155\textwidth,height=.12\textheight]{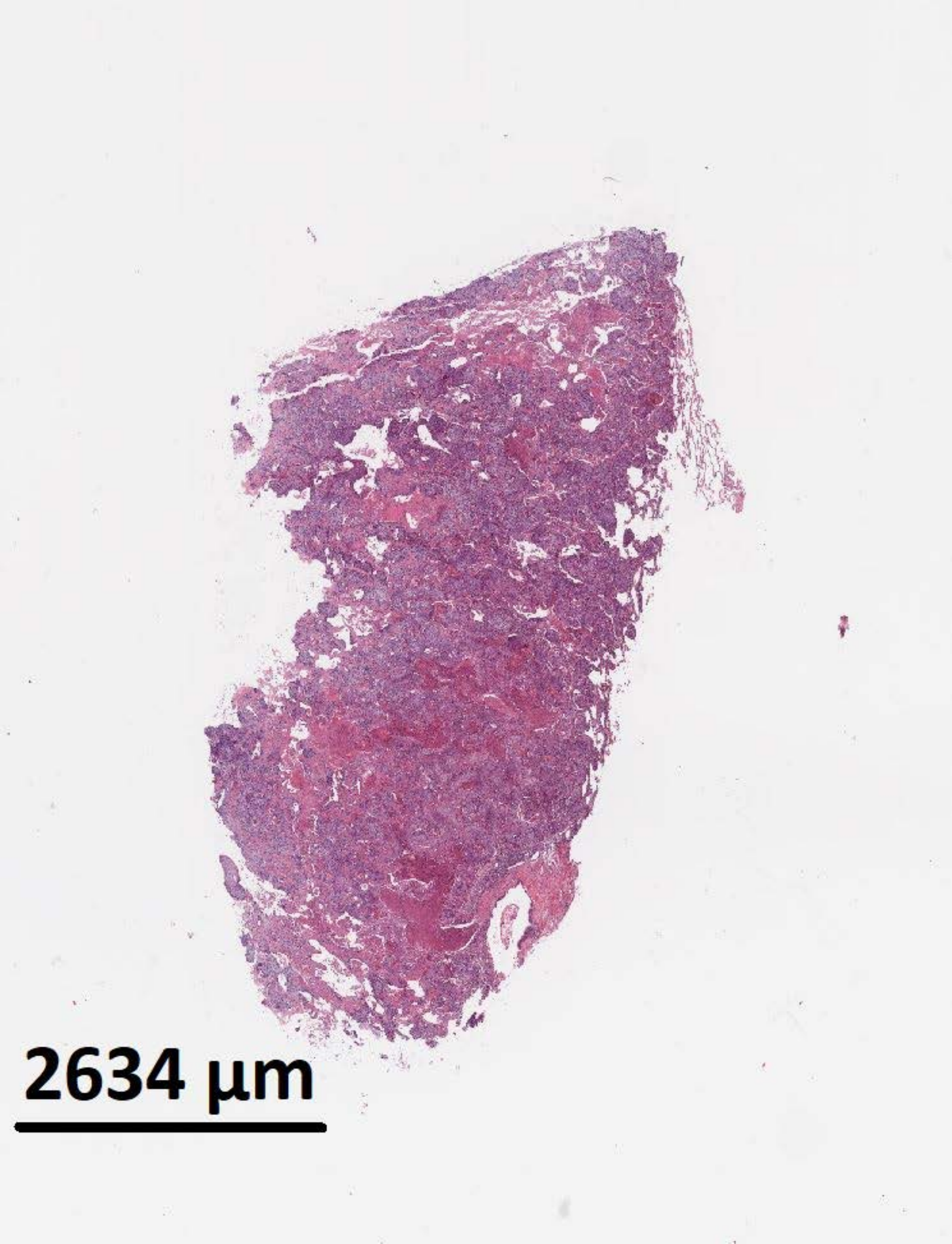} &
      \includegraphics[width=.155\textwidth,height=.12\textheight]{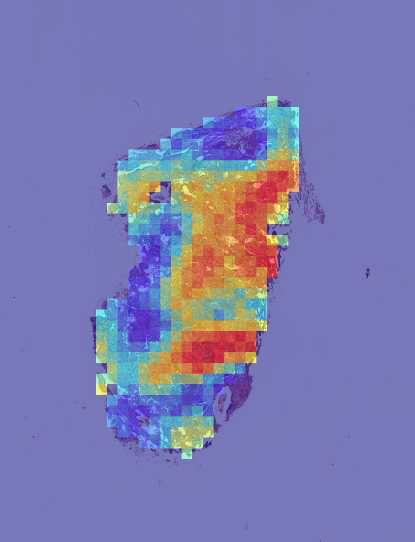} &
      \includegraphics[width=.155\textwidth,height=.12\textheight]{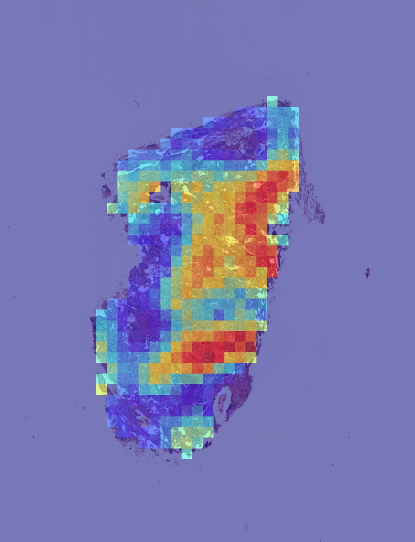} &
      \includegraphics[width=.155\textwidth,height=.12\textheight]{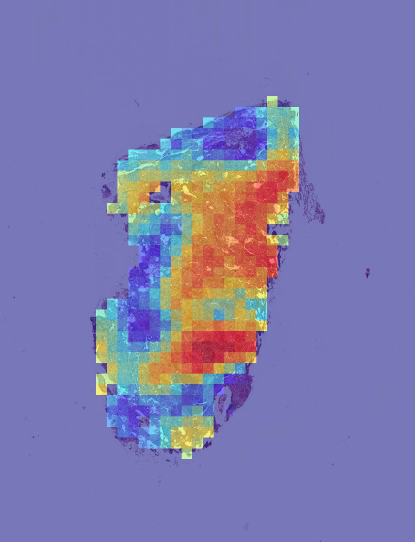} &
      \includegraphics[width=.155\textwidth,height=.12\textheight]{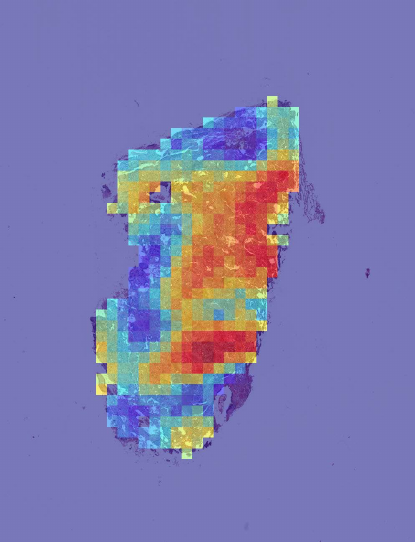} &
      \includegraphics[width=.155\textwidth,height=.12\textheight]{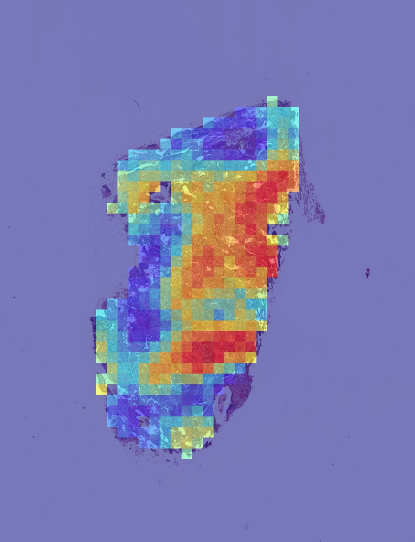} \\
      LSCC & Model 1 (p=0.94) & Model 2 (p=0.98) & Model 3 (p=0.96) & Model 4 (p=0.97) & Model 5 (p=0.98)\\
      \hline
      & \multicolumn{5}{r|}{\includegraphics[width=.06\textheight, height=0.02\textheight]{figures/colorbar_v.pdf}} \\
      \hline
    \end{tabular}
    \caption{\textbf{Graph class activation map performance.} GraphCAMs generated on WSIs across the runs performed via 5-fold cross validation are shown. The first column shows the original WSIs and the other columns show the GraphCAMs with prediction probabilities on the cross-validated model runs. The first row shows a sample WSI from the LUAD class and the second row shows an WSI from the LSCC class. The colormap represents the probability by which an WSI region is associated with the output label of interest.}
    \label{fig:more}
\vspace{-1mm}
\end{figure*}

\begin{figure}[htp]
\centering
\setlength\tabcolsep{2pt} 
\begin{tabular}{|c|cc|}
\hline
\bf{Whole Slide Image} & \multicolumn{2}{c|}{\bf{GraphCAM}} \\
\includegraphics[width=.15\textwidth, height=0.115\textwidth]{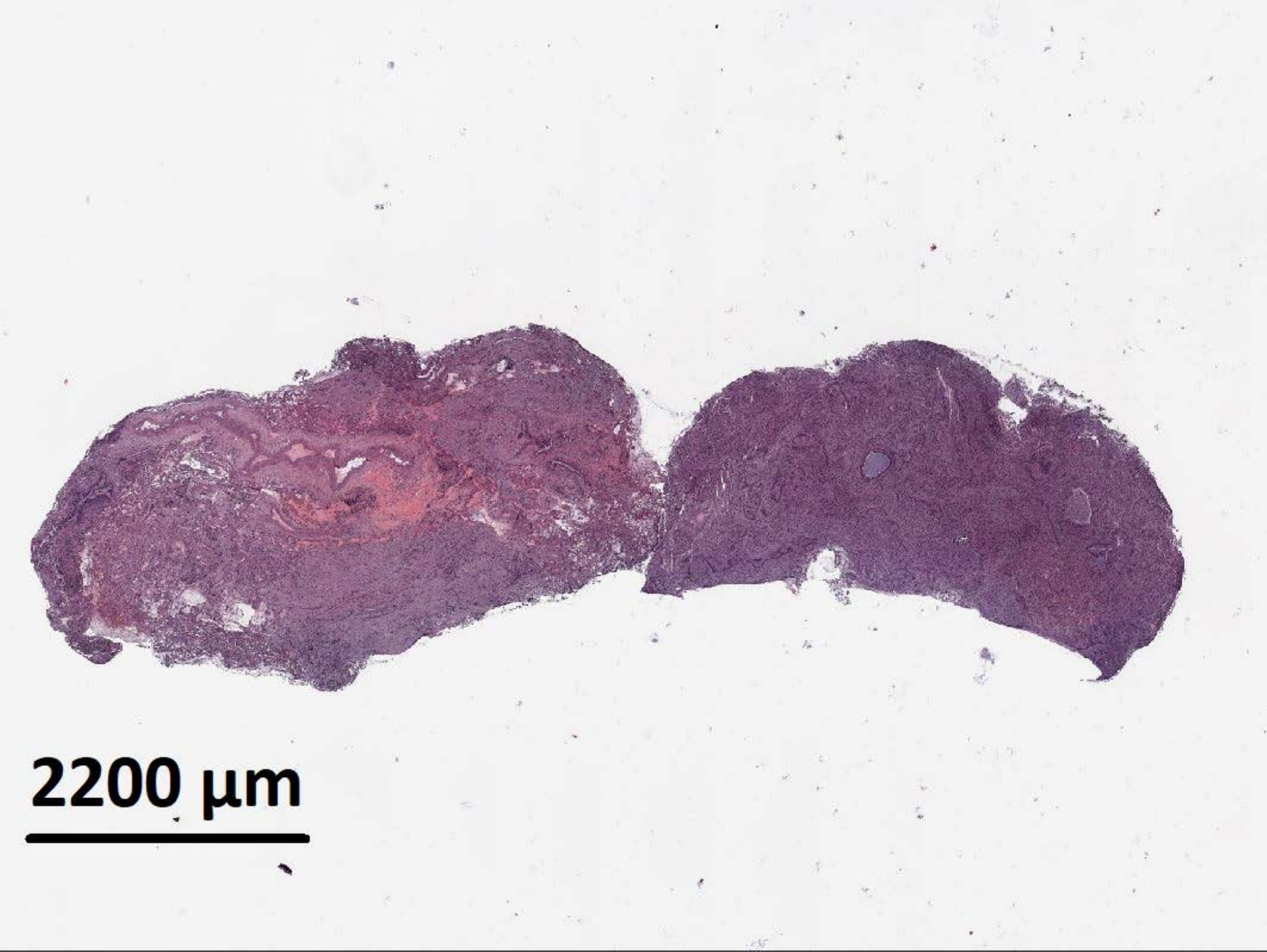} &
\includegraphics[width=.15\textwidth, height=0.115\textwidth]{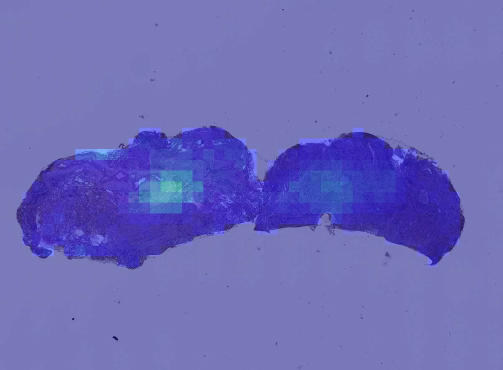} &
\includegraphics[width=.15\textwidth, height=0.115\textwidth]{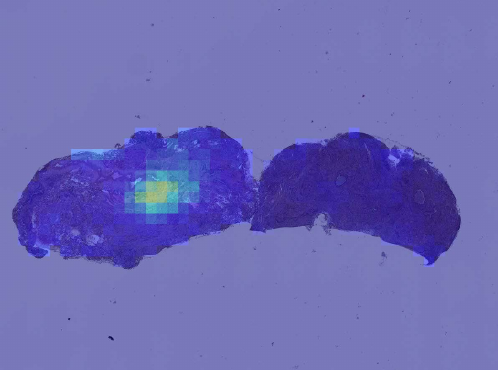} \\
LUAD & LUAD (p=0.31) & \bf{LSCC (p=0.50)} \\

\includegraphics[width=.15\textwidth, height=0.115\textwidth]{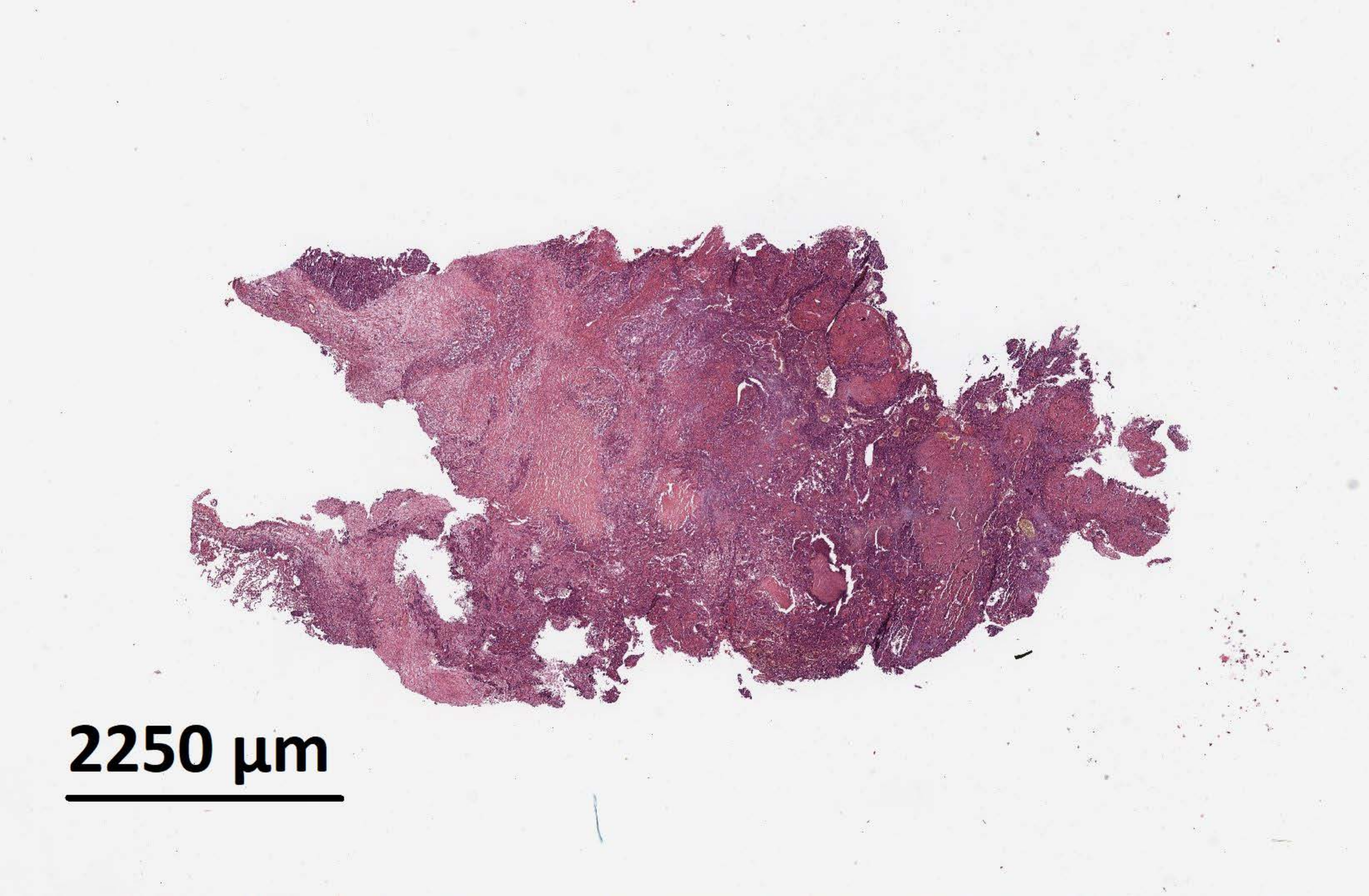} &
\includegraphics[width=.15\textwidth, height=0.115\textwidth]{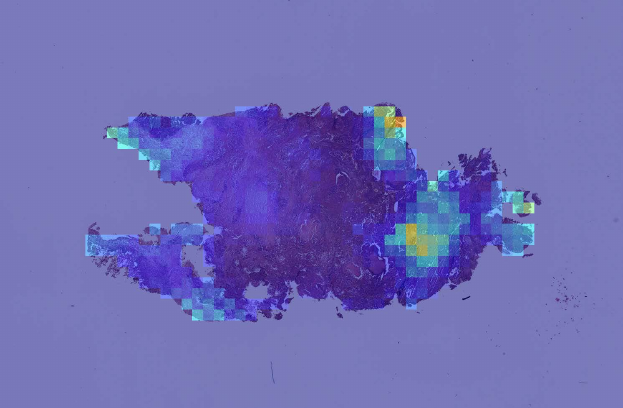} &
\includegraphics[width=.15\textwidth, height=0.115\textwidth]{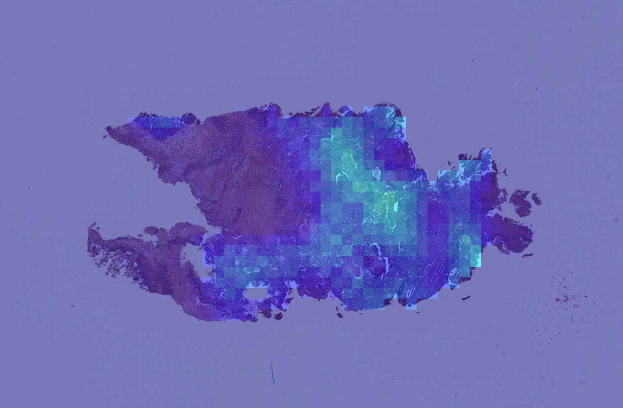} \\
LSCC & \bf{LUAD (p=0.71)} & LSCC (p=0.27) \\ \hline
  & \multicolumn{2}{r|}{\includegraphics[width=.06\textheight, height=0.02\textheight]{figures/colorbar_v.pdf}} \\
\hline
\end{tabular}
\caption{\textbf{GraphCAMs for failure cases.} The first row shows a sample WSI from the LUAD class where the model prediction was LSCC, and the second row shows an WSI from the LSCC class where the model prediction was LUAD. The first column shows the original WSI, and the second and third columns show the generated GraphCAMs along with prediction probabilities. The bold font underneath certain GraphCAMs was used to indicate the model predicted class label for the respective cases. Since this is a 3-label classification task (normal vs. LUAD vs. LSCC), the LUAD and LSCC probability values do not add up to 1.}
\label{fig:failed_cases}
\vspace{-4mm}
\end{figure}

\section{Experiments}
We performed several experiments to evaluate our GTP framework. The NLST data {($\approx 1.8$ million patches)} was exclusively used for contrastive learning to generate patch-specific features, which were used to represent each node. The GTP framework was trained on the CPTAC data ($2,071$ WSIs) using $5$-fold cross validation, and the TCGA data ($2,082$ WSIs) was used as an independent dataset for model testing using the same hyperparameters. We also conducted ablation studies to evaluate the contribution of various components on the overall GTP framework. By blocking out the GTP components, we were left with a framework that is comparable to the state-of-the-art in the field. Finally, we used GraphCAMs to identify salient regions on the WSIs, and explored their validity in terms of highlighting the histopathologic regions of interest.

\subsection{Experimental settings} \label{section:exp_setting}
Each WSI was cropped to create a bag of $512 \times 512$ non-overlapping patches at $20\times$ magnification, and background patches with non-tissue area $> 50\%$ were discarded. We used Resnet18 as the CNN backbone for the feature extractor. We adapted the Adam optimizer with an initial learning rate of $0.0001$, a cosine annealing scheme for learning rate scheduling, and a mini-batch size of $512$. We kept the trained feature extractor and used it to build the graphs. We used one graph convolutional layer, and set the transformer layer configurations as \textit{L}=$3$, MLP \textit{size}=$128$, \textit{D}=$64$ and \textit{k}=$8$ (Eq.\ref{eq:3}, Eq.\ref{eq:6}). The GTP model was trained in batches of 8 examples for $150$ iterations. The learning rate was set to $10^{-3}$ initially, and decayed to $10^{-4}$ and $10^{-5}$ at step $30$ and $100$, respectively.

\subsection{Ablation studies} \label{section:ablation}
To evaluate the effectiveness of our proposed GTP framework, we performed several ablation studies as well as compared the GTP model performance with other state-of-the-art methods (Table~\ref{table:metrics}). Since our framework leverages graph-based learning and transformer architectures, we selected a graph-based method called AttPool \cite{huang2019attpool} and a transformer-based framework called TransMIL \cite{transmil} for comparison, as they both retain spatial information to make a WSI-level prediction. To make a fair comparison, we used the same contrastive learning based model as the feature extractor for all methods. When implementing AttPool and TransMIL, we fine-tuned the hyperparameters used in the previously published original work to achieve the best performance on CPTAC and TCGA cohorts. Later, we removed the transformer component and trained the graph and compared it with the full GTP framework to study the performance of the graph-based portion of our model. Due to computational constraints, it was not feasible to remove the graph component and train the transformer; there was an out-of-memory issue when the number of patches exceeded $4,760$ for batch size of $1$ on a GeForce RTX 2080 Ti, GPU memory: 11Gb workstation. 

We also performed a detailed analysis to evaluate the effect of contrastive learning on the GTP model performance by conducting studies with and without it (Table~\ref{table:ablation}). We adopted the supervised learning-based model, Resnet \cite{He2016DeepRL}, instead of contrastive learning-based model. Resnet$^{\star}$ represents the model trained on ImageNet \cite{deng5206848}. Resnet$^{\dagger}$ represents Resnet$^{\star}$ fine-tuned on the same NLST data which is used for contrastive learning. To train Resnet$^{\dagger}$ via supervised learning, we assigned the WSI label to all patches belonging to the same WSI. We also added an unsupervised learning model, a convolutional autoencoder (CAE) \cite{cae}, for comparison. Additionally, to make a fair comparison in terms of domain specificity, we added the contrastive learning model trained on STL10 \cite{stl10} to compare to the model trained on NLST. The STL-10 dataset is an image recognition dataset for developing unsupervised feature learning, whose images were acquired from labeled examples on ImageNet, whereas the NLST dataset is a lung cancer imaging dataset. In essence, these ablation studies allowed us to fully evaluate the power of the GTP framework for WSI-level classification.

\subsection{Computing infrastructure}
We used PyTorch (v1.9.0) and a NVIDIA 2080Ti graphics card with 11 GB memory on a GPU workstation to implement the model. The training speed was about $2.4$ iterations/s, and it took less than a day to reach convergence. The inference speed was $<1$~s per WSI with a batch size of $2$. 

\subsection{Performance metrics}
For the classification task (LUAD vs.\ LSCC vs.\ normal), we generated receiver operating characteristic (ROC) and precision-recall (PR) curves based on model predictions on the CPTAC testing and the full TCGA datasets. For each ROC and PR curve, we also computed the area under curve (AUC), precision, recall, specificity, and accuracy. Since we used 5-fold cross validation, we took all the curves from different folds and calculated the mean AUCs and their variance. Delong's statistical test was used to show whether the AUCs of two different models were significantly different. GraphCAMs were used to generate visualizations and gain a qualitative understanding on the model performance.

\subsection{Expert annotations}
A randomly selected set of WSIs (n=10 for LUAD and n=10 for LSCC) were uploaded to a secure, web-based software (PixelView; deepPath, Boston, MA) for expert review. Using an Apple Pencil and an iPad, E.J.B. annotated several histologic features of LUAD and LSCC.  Tumor regions of LUAD were annotated by their LUAD tumor subtypes (solid, micropapillary, cribiform, papillary, acinar, and lepidic).  For both LUAD and LSCC, other histologic features of the tumor were also annotated including necrosis, lymphatic invasion, and vascular invasion.  Non-tumor regions were annotated as normal or premalignant epithelium, normal or inflammed lung, stroma, and cartilage.  The annotations of each region are reflective of the most predominant histologic feature.  The annotations were grouped into tumor and non-tumor categories and exported as binary images and processed to quantify the extent of overlap between the model-derived GraphCAMs and the pathologist-driven annotations. Intersection over union (IoU) was used as the metric to measure the overlap between the model and the expert.

\begin{table*}[t]
\caption{\textbf{Ablation studies on the feature extractors in 3-label (Normal vs. LUAD vs. LSCC)) classification task.} We used different feature extractors for graph construction and evaluated their role on the overall classification task. Here,  Resnet$^{\star}$ indicates the use of a pre-trained Resnet18 network without fine-tuning, Resnet$^{\dagger}$ with fine-tuning.
CAE is a convolutional auto encoder. CL represents contrastive learning on STL10 or NLST by our method. Mean performance
metrics are reported along with the corresponding values of standard deviation in parentheses.
}

\centering
\setlength\tabcolsep{2pt} 
\subfloat[Precision, Recall/Sensitivity, and Specificity (Percentage (\%) values are reported).]{
\begin{tabular}{c|c|ccc|ccc|ccc}
\hline
\multirow{2}{*}{Method} & \multirow{2}{*}{Data} & \multicolumn{3}{c|}{Precision} & \multicolumn{3}{c|}{Recall/Sensitivity}  & \multicolumn{3}{c}{Specificity}  \\
 & & Normal     & LUAD        & LSCC        & Normal     & LUAD        & LSCC        & Normal     & LUAD        & LSCC  \\ \hline
\multicolumn{1}{c|}{\multirow{2}{*}{Resnet$^{\star}$}} & CPTAC  & $88.3(4.2)$ & $63.8(2.7)$ & $76.4(5.0)$ & $81.1(5.9)$ & $71.0(4.1)$ & $73.6(7.6)$ & $94.0(3.1)$ & $80.9(1.8)$ & $88.4(3.6)$  \\
\multicolumn{1}{c|}{}                                  & TCGA  & $56.2(7.2)$ & $42.6(3.1)$ & $35.6(10.3)$ & $50.9(11.3)$ & $31.3(24.4)$ & $50.1(22.8)$ & $79.7(8.6)$ & $77.7(19.8)$ & $58.3(11.3)$ \\ \hline

\multicolumn{1}{c|}{\multirow{2}{*}{Resnet$^{\dagger}$}}  & CPTAC  & $88.1(5.1)$ & $77.6(5.6)$ & $77.7(4.1)$ & $89.4(3.5)$ & $72.2(6.6)$ & $80.1(6.6)$ & $93.3(3.6)$ & $89.6(3.8)$ & $88.3(3.6)$ \\
\multicolumn{1}{c|}{}                                     & TCGA  & $62.1(2.1)$ & $51.8(3.6)$ & $66.6(5.6)$ & $77.8(3.2)$ & $49.5(9.3)$ & $52.5(6.4)$ & $77.6(2.9)$ & $76.3(5.9)$ & $85.6(4.8)$ \\ \hline

\multicolumn{1}{c|}{\multirow{2}{*}{CAE}}  & CPTAC  & $90.6(1.7)$ & $77.8(2.2)$ & $73.2(2.7)$ & $88.5(4.2)$ & $65.6(4.5)$ & $85.8(3.6)$ & $95.1(1.0)$ & $91.1(1.2)$ & $84.3(2.9)$  \\
\multicolumn{1}{c|}{}                      & TCGA  & $61.6(2.3)$ & $42.3(2.5)$ & $57.2(2.1)$ & $78.3(3.2)$ & $35.9(1.6)$ & $51.1(5.7)$ & $76.9(2.7)$ & $75.1(2.1)$ & $80.0(1.8)$ \\ \hline

\multirow{2}{*}{\begin{tabular}[c]{@{}c@{}}CL\\ (STL10)\end{tabular}}  & CPTAC  & $\mathbf{95.3(1.6)}$ & $87.6(3.5)$ & $\mathbf{91.7(3.4)}$ & $\mathbf{96.2(1.6)}$ & $\mathbf{90.5(4.2)}$ & $87.3(3.7)$ & $\mathbf{97.5(0.9)}$ & $93.8(2.2)$ & $\mathbf{95.9(1.8)}$  \\
\multicolumn{1}{c|}{}                                                  & TCGA  & $76.9(2.7)$ & $66.5(3.3)$ & $73.8(3.5)$ & $82.2(5.5)$ & $73.4(1.5)$ & $61.0(4.1)$ & $88.4(1.8)$ & $81.0(3.1)$ & $88.5(2.8)$ \\ \hline

\multirow{2}{*}{\begin{tabular}[c]{@{}c@{}}CL\\ (NLST)\end{tabular}}  & CPTAC  & $93.2(3.0)$ & $\mathbf{88.4(3.9)}$ & $87.8(3.0)$ & $95.9(2.2)$ & $83.9(4.5)$ & $\mathbf{89.2(4.0)}$ & $96.2(1.7)$ & $\mathbf{94.7(1.9)}$ & $93.8(1.7)$  \\
\multicolumn{1}{c|}{}                                                 & TCGA  & $\mathbf{89.2(2.8)}$ & $\mathbf{74.4(2.7)}$ & $\mathbf{84.4(0.7)}$ & $\mathbf{92.6(2.7)}$ & $\mathbf{79.8(1.9)}$ & $\mathbf{75.2(1.6)}$ & $\mathbf{94.7(1.6)}$ & $\mathbf{86.0(2.3})$ & $\mathbf{92.7(0.4)}$ \\ \hline

\end{tabular}}

\vspace{0.3cm}

\centering
\setlength\tabcolsep{5pt} 
    \begin{subtable}{.5\linewidth}
      \centering
        \caption{Accuracy and AUC (Percentage (\%) values are reported).}

        \begin{tabular}{c|c|c|c}
        \hline
        \multicolumn{1}{c|}{Method} & \multicolumn{1}{c|}{Data} & \multicolumn{1}{c|}{Accuracy} & \multicolumn{1}{c}{AUC} \\ \hline
        \multirow{2}{*}{Resnet$^{\star}$}  & CPTAC                        & $75.4(2.0)$ & $88.9(0.4)$   \\
                                           & TCGA                        & $44.0(2.8)$ & $60.4(3.0)$   \\ \hline
        \multirow{2}{*}{Resnet$^{\dagger}$}   & CPTAC                     & $80.8(1.1)$ & $92.2(0.9)$   \\
                                     & TCGA                              & $59.6(1.1)$ & $76.2(0.8)$   \\ \hline
        \multirow{2}{*}{CAE}   & CPTAC                                    & $80.2(1.6)$ & $93.2(0.9)$   \\
                                     & TCGA                              & $54.7(2.2)$ & $72.9(1.6)$   \\ \hline
        \multirow{2}{*}{\begin{tabular}[c]{@{}c@{}}CL\\ (STL10)\end{tabular}}   & CPTAC    & $\mathbf{91.4(1.1)}$ & $\mathbf{98.0(0.9)}$   \\
                                   & TCGA                                & $71.9(1.9)$ & $86.4(1.6)$   \\ \hline 
        \multirow{2}{*}{\begin{tabular}[c]{@{}c@{}}CL\\ (NLST)\end{tabular}}    & CPTAC    & $91.2(2.5)$ & $97.7(0.9)$   \\
                                   & TCGA                                & $\mathbf{82.3(1.0)}$ & $\mathbf{92.8(0.3)}$   \\ \hline 
        \end{tabular}
    \end{subtable}%
    \begin{subtable}{.5\linewidth}
      \centering
        \caption{DeLong’s test to compare the AUCs of models generated using different feature extractors. $\text{log}_{10}(0.05)=-1.301$.}
        \begin{tabular}{c|c|c}
        \hline
        \multicolumn{1}{c|}{Method} & \multicolumn{1}{c|}{Data} & \multicolumn{1}{c}{$\text{log}_{10}$(p-value)}  \\ \hline
        \multirow{2}{*}{Resnet$^{\star}$}  & CPTAC                        & $-3.711(2.044)$     \\
                                           & TCGA                        & $-8.382(0.992)$     \\ \hline
        \multirow{2}{*}{Resnet$^{\dagger}$}   & CPTAC                     & $-2.336(0.592)$     \\
                                     & TCGA                              & $-4.041(0.418)$     \\ \hline
        \multirow{2}{*}{CAE}   & CPTAC                                    & $-2.325(2.573)$    \\
                                     & TCGA                              & $-4.532(0.652)$     \\ \hline
        \multirow{2}{*}{\begin{tabular}[c]{@{}c@{}}CL\\ (STL10)\end{tabular}}   & CPTAC    & $-1.773(1.397)$     \\
                                   & TCGA                          & $-2.962(1.482)$     \\ \hline 
        \end{tabular}
    \end{subtable}

\label{table:ablation}
\end{table*}

\begin{table}[t]
\caption{\textbf{Ablation studies on model hyperparameters.} Various hyperparameters were varied to evaluate their effect on the overall model performance. All the models are trained and evaluated using a portion of data from the CPTAC cohort, and model accuracy was reported on the left-out CPTAC cases (last column). The MLP dimension of the model was 128. We used WSIs with 20x magnification for all cases. We used non overlapping patches for all cases except for ${\star}$. The batch size used was 8 except for ${\dagger}$. For all these studies, the CPTAC data was randomly divided in $7:3$ ratio, where $70\%$ data was used for training and the rest for testing. }
\setlength\tabcolsep{3pt}
\begin{adjustbox}{width=\columnwidth,center}
\begin{tabular}{cccc|lc|l}
\hline
\multicolumn{4}{c|}{Model configuration}                                                                                                                                                                                                                                                                         & \multicolumn{2}{c|}{Graph configuration}                                                                                                                & \multirow{3}{*}{Accuracy} \\ \cline{1-6}
\multirow{2}{*}{\begin{tabular}[c]{@{}c@{}}Hidden \\ dimension\end{tabular}} & \multirow{2}{*}{\begin{tabular}[c]{@{}c@{}}GCN \\ layer\end{tabular}} & \multirow{2}{*}{\begin{tabular}[c]{@{}c@{}}Transformer \\ block\end{tabular}} & \multirow{2}{*}{\begin{tabular}[c]{@{}c@{}}min-cut \\ node\end{tabular}} & \multirow{2}{*}{\begin{tabular}[c]{@{}c@{}}patch \\ size\end{tabular}} & \multirow{2}{*}{\begin{tabular}[c]{@{}c@{}}node\\ connectivity\end{tabular}} &                           \\
                                                                             &                                                                       &                                                                               &                                                                          &                                                                        &                                                                               &                           \\ \hline
128                                                                          & 3                                                                     & 3                                                                             & 120                                                                      & 512                                                                    & 8                                                                             & \bf{0.925} \\ \hline
128 & 3 & 3 & 100 & 512 & 8   & 0.915 \\
128 & 3 & 3 & 80 & 512 & 8   & 0.903 \\   
128 & 1 & 3 & 100 & 512 & 8  & 0.908 \\   
128 & 3 & 6 & 100 & 512 & 8  & 0.894 \\
128 & 1 & 6 & 100 & 512 & 8  & 0.919 \\
128 & 1 & 6 & 80  & 512 & 8  & 0.906 \\
128 & 1 & 6 & 120 & 512 & 8  & 0.898 \\
128 & 1 & 3 & 120 & 512 & 8  & 0.911 \\ 
64  & 3 & 3 & 120 & 512 & 8  & 0.908 \\
64  & 1 & 3 & 100 & 512 & 8  & 0.903 \\ 
64  & 1 & 6 & 100 & 512 & 8  & 0.913 \\ 
64  & 3 & 3 & 100 & 512 & 8  & 0.901 \\ 
64  & 3 & 6 & 100 & 512 & 8  & 0.901 \\ \hline
128 & 3 & 3 & 120 & 512 & 4   & 0.896  \\
128 & 3 & 3 & 120 & $512^{\star}$ & 8 & 0.881  \\
128 & 3 & 3 & 120 & 1024 & 8   & 0.864 \\
128 & 3 & 3 & 120 & 360 & 8   & $0.898^{\dagger}$ \\ \hline
\end{tabular}
\end{adjustbox}
\begin{tablenotes}
\item[${\star}$] patches have overlap (10\%). \quad \item[${\dagger}$] Batch size is 2 due to memory limitation. \\
\end{tablenotes}
\label{table:params}
\end{table}

\section{Results} \label{section:results}
The GTP framework that leveraged contrastive learning followed by fusion of a graph with a transformer provided accurate prediction of WSI-level class label (Table~\ref{table:metrics}). High model performance was observed on the normal vs.\ LUAD vs.\ LSCC task on the CPTAC test dataset but dropped slightly on the TCGA dataset. High model performance was also confirmed via the ROC and PR curves generated on both the CPTAC and TCGA datasets for all the classification tasks (Fig.~\ref{fig:roc_pr_3_label}). On each task, the mean area under the ROC and PR curves was high (all $>0.9$) on the CPTAC test data. For the TCGA dataset, which was used for external testing, the mean area under the ROC and PR curves dropped slightly, especially for the LUAD and LSCC classification tasks. The model leaned towards incorrectly classifying a few LSCC and LUAD cases but correctly classified most of the WSIs with no tumor. 

The GTP framework achieved the best performance compared with other state-of-the-art methods such as TransMIL \cite{transmil} or AttPool \cite{huang2019attpool} (Table~\ref{table:metrics}). The AttPool framework selects the most significant nodes in the graph and aggregates information via the attention mechanism. The transMIL framework uses a self-attention mechanism from the transformer to model the interactions between all patches from the WSI for information aggregation. We submit that the graph structure along with graph pooling enabled the GTP to capture the short-range associations while the transformer helped capture the long-range associations. Also, since we were able to perform batch training instead of just a single sample per iteration, the model achieved faster convergence and also processed any potential variabilities between multiple samples in the training process.

The GT-based class activation maps (GraphCAMs) identified WSI regions that were highly associated with the output class label (Fig.~\ref{fig:cam}). We also observed a high degree of overlap between the expert-identified regions of interest with the binarized GraphCAMs. Specifically, the maximum value of the intersection over union (IoU) was $0.857$ at threshold probability of $0.6$ for the LUAD case (Row 1 in Fig.~\ref{fig:cam}), and an IoU of $0.733$ at threshold probability of $0.2$ for the LSCC case (Row 2 in Fig.~\ref{fig:cam}). Additionally, the expert pathologist selected $20$ cases from the TCGA cohort ($10$ LUAD and $10$ LSCC) and annotated the tumor regions on them. We then estimated the overlap between the model-identified regions of interest (via GraphCAMs) and the pathologist annotations, resulting in a high degree of agreement (Mean of maximum values of IoU $=0.817\pm0.165$). Importantly, the degree of overlap remained fairly consistent as the threshold for binarization varied, supporting that our approach to identifying important WSI regions has clinicopathologic significance. We also observed that the same set of WSI regions were highlighted by our method across the various cross-validation folds (Fig.~\ref{fig:more}), thus indicating consistency with our technique in identifying salient regions of interest. Also, since we can generate class-specific probability for each GraphCAM, our approach allows for better appreciation of the model performance and its interpretability in predicting an output class label. We must however note that in certain cases when the model fails to predict the class label, the GraphCAMs may not result in interpretable findings (Fig.~\ref{fig:failed_cases}).

Ablation studies revealed that our GTP framework that used contrastive learning and combined a graph with a transformer served as a superior model for WSI-level classification (Table~\ref{table:ablation}). When contrastive learning was replaced with a pre-trained architecture (Resnet18 with and without fine tuning), the model performance for the 3-label classification task dropped. The reduction in performance was evident on both CPTAC and TCGA datasets. The model performance also dropped when we trained a novel convolutional auto-encoder \cite{cae} in lieu of contrastive learning. These results imply that the feature maps generated via contrastive learning were informative to encode a variety of visual information for GT-based classification with a fair degree of generalizability. We also studied the effect of the domain specificity of the dataset on the model performance, by comparing contrastive learning models trained on STL10 and NLST (Table~\ref{table:ablation}). The STL10\cite{stl10} is an image recognition dataset for developing unsupervised feature learning. The images are acquired from labeled examples on ImageNet. The contrasive learning models trained on NLST performs significantly better than STL10, indicating that the domain knowledge learned by contrastive learning helps improve the generalizability of our GTP model. (Table~\ref{table:ablation}). 

To analyze the impact of different model hyperparameters and graph configurations on the classification performance, we performed a series of ablation studies (Table~\ref{table:params}). In general, the consistency of the results generated using different choices of hyperparameters demonstrate robustness of our proposed model. When the dimension of the hidden state was reduced by half (i.e., $128$ to $64$), we noticed that the model performance degraded. Similar results in model performance were observed when the number of GCN layers reduced from $3$ to $1$. In essence, additional GCN layers increase the receptive field of the nodes and could continue to provide long-term information during model training. Increasing the number of min-cut nodes improved the performance, since typically representing the data using more tokens can contribute to higher prediction accuracy \cite{dosovitskiy2020vit}. Also, we found that $3$ transformer blocks were sufficient to integrate information across the entire pooled nodes from the graph. It must be noted that using more parameters for model training might alter the performance but would be feasible only at a higher computational cost. For instance, the cost can increase quadratically with respect to the token number in the self-attention blocks. Nevertheless, to find a trade-off between computational efficiency and model effectiveness, we tested various parameters including the effect of different forms of the graph construction. Increasing the patch size lowered the accuracy. Using 4-node connectivity instead of 8-node connectivity reduced the receptive field, resulting in lower accuracy. Choosing smaller patch sizes increased the size of graph, which increased the computational cost. Lastly, when a $10\%$ overlap between patches was considered and graph constructed, the model performance slightly reduced. These findings indicate that our proposed GTP framework along with the selected hyperparameters is capable of reasonably integrating information across the entire WSI to predict WSI-level output of interest.

\section{Discussion}
In this work, we developed a deep learning framework that integrates graphs with vision transformers to generate an efficient classifier to differentiate normal WSIs from those with LUAD or LSCC. Based on the standards of various model performance metrics, our approach resulted in classification performance that exceeded other deep learning architectures that incorporated various state-of-the-art configurations. Also, our novel class activation mapping technique allowed us to identify salient WSI regions that were highly associated with the output class label of interest. Finally, we found that our model-identified regions of interest closely matched with pathologist-derived assessments on the WSIs. Thus, our findings represent novel contributions to the field of interpretable deep learning while also simultaneously advancing the fields of computer vision and digital pathology.

The field of computational pathology has made great strides recently due to advancements in vision-based deep learning. Still, owing to the sheer size of pathology images generated at high resolution, WSI assessment that can integrate spatial signatures along with local, region-specific information for prediction of tumor grade remains a challenge. A large body of work has focused on patch-level models that may accurately predict tumor grade but fail to capture spatial connectivity information. As a result, identification of important image-level features via such techniques may lead to inconsistent results. Our GTP framework precisely tackled this scenario by integrating WSI-level information via a graph structure and thus represents a key advancement in the field. 

One of the novel contributions in our work is the generation of graph-based class activation maps (GraphCAM), which can highlight the WSI regions that are associated with the output class label. Unlike other saliency mapping techniques such as attention rollout \cite{rollout} or layer-wise relevance propagation \cite{Binder2016LayerWiseRP}, GraphCAMs can generate class-specific heatmaps. This is a major advantage because an image may contain information pertaining to multiple classes, and for these scenarios, identification of class-specific feature maps becomes important. This is especially true in real-world scenarios such as pathology images containing lung tumors. Class-specific features could be useful, for example, in identifying rare mixed histology adenosquamous tumors that contain both LUAD and LSCC patterns \cite{NICHOLSON2022362}. In such cases, training well-known supervised deep learning classifiers such as convolutional neural networks that use the overall WSI label for classification at patch-level or even at the WSI-level may not necessarily perform well and even misidentify the regions of interest associated with the class label. By generating class-specific CAMs learned at the WSI-level, our GTP approach provides a better way to identify regions of interest on WSIs that are highly associated with the corresponding class label. 

Our study has a few limitations. We leveraged contrastive learning to generate patch-level feature vectors before constructing the graph, which turned out to be a computationally intensive task. Future studies can explore other possible techniques for feature extraction that can lead to improved model performance. Our graph was constructed by dividing the WSI into image patches, followed by creation of nodes using the embedding features from these patches. Other alternative ways can be explored to define the nodes and create graphs that are more congruent and spatially connected. While we have demonstrated the applicability of GTP to lung tumors, extension of this framework to other cancers is needed to fully appreciate its role in terms of assessing WSI-level correlates of disease. In fact, our method is not specific to cancers and could be adapted to other computational pathology applications.

In conclusion, our GTP framework produced an accurate, computationally efficient model by capturing regional and WSI-level information to predict the output class label. GTP can tackle high resolution WSIs and predict multiple class labels, leading to generation of interpretable findings that are class-specific. Our GTP framework could be scaled to WSI-level classification tasks on other organ systems and has the potential to predict response to therapy, cancer recurrence and patient survival.

\bibliographystyle{abbrv}
\bibliography{refs}
\end{document}